\documentclass{article} %
\usepackage{iclr2027_conference,times}

\usepackage{amsmath,amsfonts,bm}

\def\eqref#1{equation~\ref{#1}}

\def\1{\bm{1}}

\DeclareMathAlphabet{\mathsfit}{\encodingdefault}{\sfdefault}{m}{sl}
\SetMathAlphabet{\mathsfit}{bold}{\encodingdefault}{\sfdefault}{bx}{n}

\usepackage{hyperref}
\usepackage{url}
\usepackage{graphicx}
\usepackage{booktabs}
\usepackage{xcolor}
\usepackage{adjustbox}
\usepackage{placeins}

\newcommand{\figorbox}[3]{\IfFileExists{#1}{\includegraphics[width=#2]{#1}}{\fbox{\parbox[c][#3][c]{#2}{\centering draft: \texttt{#1}}}}}

\title{Break Step: Recursive Training Resonates\\with Replayed Sampling Noise}

\author{Yangze Liu \\
Shandong University \\
\texttt{yangze2@illinois.edu}
\And
Zhongyi Han\thanks{Corresponding author.} \\
Shandong University \\
\texttt{zhongyi.han@sdu.edu.cn}}

\iclrfinalcopy

\begin{document}

\maketitle

\begin{abstract}
How fast does a language model degrade when trained on its own outputs? Theory traces it to gradually accumulating errors, while experiments report repeated phrases within ten generations. Under a fixed sampling seed in vLLM, the fast loss of lexical diversity comes from the sampler. When vLLM serves a batch from one seeded sampling configuration, every request receives the same random draws, and a fixed seed replays them every generation. Fine-tuning raises the tokens that won, and the replayed draws let them win by more. Sharing across requests and replay across generations matter only together. Remove either one, by changing the shared seed every generation or by giving each request its own seed that repeats every generation, and the unique-4-gram fraction of two StableLM checkpoints stays near its starting value of about 0.98 through generation 3. Keep both, and the replayed shared seed takes seven checkpoints from five families to between 0.045 and 0.38 by then. Three generations of replay write the favoured phrases into the weights: decoded with one seed per request, the generation-3 weights of the replayed StableLM-2-1.6B chain recover most of their diversity, yet the phrase that filled every sample under the shared seed still opens 46\% of them. Without replay, five checkpoints drift slowly, consistent with the gradual accumulation that theory describes, and three turn incoherent though their diversity scores stay high. One peer-reviewed model-collapse pipeline that fine-tunes Gemma-2-27B samples identical prompts under one seeded configuration, and three quarters of the rows it released for one iteration repeat nearly as often as one such batch copies them. A seed per request restores the fresh sample that stability analyses assume.
\end{abstract}

\section{Introduction}
\label{sec:intro}

In 1831 a column of soldiers marching in step across the Broughton suspension bridge near Manchester set it swinging and brought it down, and armies since order their troops to break step on bridges. The danger is the same push arriving at the same phase, step after step, so that each swing feeds the next. This paper shows the same shape of failure in recursive training: a sampler that replays its random draws in every generation pushes the training loop at the same phase.

Training generative models on their own outputs narrows what they produce \citep{shumailov2024ai,alemohammad2024selfconsuming}, and how fast it does so is not settled. Theory locates the cause in the errors of each generation, the finite-sample error of refitting to a finite synthetic sample and the approximation error of the model class. These errors accumulate gradually with the number of generations, the finite-sample part at a rate that falls as the synthetic sample grows, and they stay bounded when the real data stay in the training set \citep{shumailov2024ai,gerstgrasser2024collapse,bertrand2024stability,dohmatob2024demystified}. Experiments that retrain language models on their own text report repeated phrases within ten generations \citep{shumailov2024ai} and outputs that shrink to a handful of distinct strings \citep{briesch2023large}. The theory treats the synthetic samples of a generation as fresh draws from the current model. In practice the samples come from an inference engine, and the engine decides how the random numbers behind them are drawn.

Two self-training runs of the same checkpoint make the point. Each run fine-tunes StableLM-2-1.6B from its base weights on 2{,}100 of its own continuations per generation for three generations, and the two differ in one line of the generation script. With \texttt{seed=42} in every generation, the fraction of unique word 4-grams falls from 0.98 to 0.15. With \texttt{seed=42+1000*g} in generation $g$, it ends at 0.95. The prompts, the fine-tuning recipe, the engine and the checkpoint are the same in both runs.

\begin{figure}[t]
\centering
\figorbox{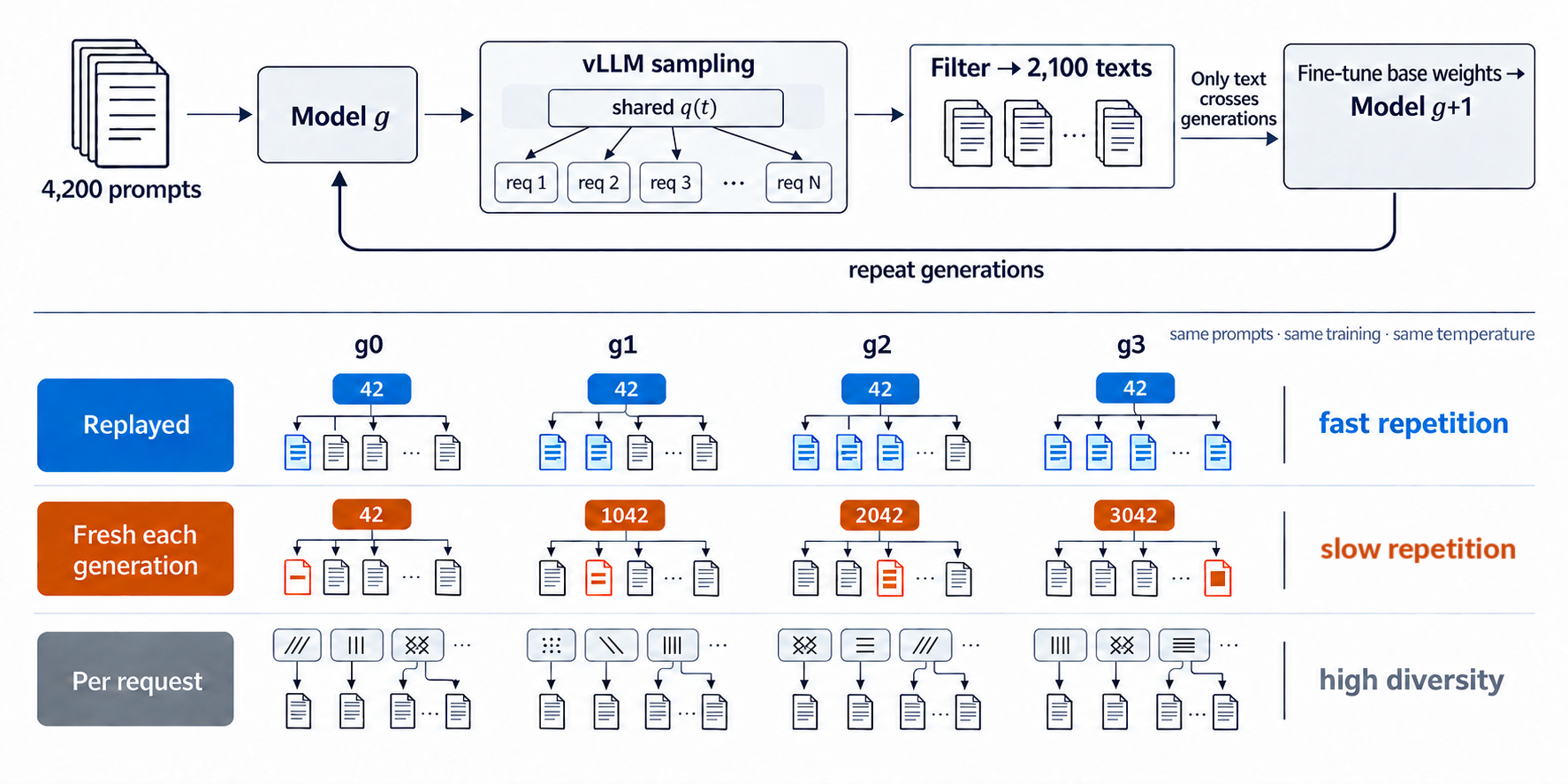}{0.86\linewidth}{2.1in}
\caption{Self-training pipeline and seed rules (schematic). Each generation continues 4{,}200 prompts, filters to 2{,}100 texts, and fine-tunes a fresh copy of the base weights on them. $q(t)$ is the vector of random draws, one per vocabulary id, that picks the token at step $t$ (\S\ref{sec:mechanism}), and a seeded batch shares it across requests. Replayed: one seed serves every request and returns each generation, so a phrase favoured by its draws grows under the loop. Fresh: the shared seed changes to $42+1000g$ in generation $g$, so each generation writes shared phrases of its own and repetition builds slowly (\S\ref{sec:resonance}). Per request: each request has its own seed. The row draws the independent rule; the fixed per-request rule, which keeps each seed across generations, is in Table~\ref{tab:twobytwo}.}
\label{fig:mechanism}
\end{figure}

The fast time scale of the first run is a resonance (\S\ref{sec:mechanism}). vLLM samples a token by drawing one exponential variate per vocabulary id and emitting the id with the largest ratio of probability to its draw. A request that carries a seed gets its own generator, so one seed for a batch hands every request the same draws at every step, a single lottery shared by the whole batch, and the same seed in the next generation hands them the same draws again. Fine-tuning on generation $g$ raises the tokens that won, and identical draws at generation $g{+}1$ make them win by more. Recursive training is a feedback loop, and the replayed draws push it at the same phase in every generation, as the soldiers' step pushed the bridge (Figure~\ref{fig:mechanism}). A new seed in every generation changes the step without breaking it: the raised tokens meet draws that favour them only by chance, so the loop slows, but the whole batch still shares each generation's draws. A seed per request breaks step.

The replayed seed does two things, and we separate them: it shares its draws across the batch, and it returns them in every generation. Four seed rules cross the two (Table~\ref{tab:twobytwo}). Independent seeds remove both and keep the diversity of all seven checkpoints that lose it under the replay (\S\ref{sec:necessary}), so recursive training alone does not produce the fast loss, and the repeated phrases are in the draws before any training (\S\ref{sec:fingerprint}). A shared seed that changes every generation keeps the sharing, a per-request seed that returns every generation keeps the return, and in the two StableLM checkpoints run under all four rules both hold lexical diversity near its starting value through generation 3 (\S\ref{sec:resonance}). The fast loss needs a shared draw that returns. It is an interaction inside the loop of sampling and training, not a cost of a fixed seed or of reproducible sampling as such. It grows with the share of requests on the replayed seed (\S\ref{sec:dose}), and after three generations the favoured phrases are in the weights: decoded with independent seeds, the replayed chain's weights still open nearly half their samples with the phrase that filled all of them under the shared seed (\S\ref{sec:xdec}). The loss does not appear at temperature 0.7, and repeated prompts alone do not start it (Appendix~\ref{sec:nontriggers}).

In these self-loops the fast loss of lexical diversity comes from this resonance. Without the replay five of the eight checkpoints drift slowly, which is consistent with the gradual accumulation of error that theory describes. Granite-3.3-2B and two Qwen checkpoints instead turn incoherent within three generations, which perplexity and embedding drift register and the 4-gram count does not (\S\ref{sec:remains}). One peer-reviewed model-collapse pipeline that fine-tunes Gemma-2-27B broadcasts one seeded configuration to a batch, and its released data hold copies of the size one such batch writes (\S\ref{sec:wild}). The remedy is one line: give each request its own seed (\S\ref{sec:discussion}).

\section{Related work}
\label{sec:related}

Recursive training on synthetic data narrows diversity for images and text \citep{shumailov2024ai,alemohammad2024selfconsuming,briesch2023large,guo2024curious}. Later work lets synthetic data accumulate with real data \citep{gerstgrasser2024collapse,kazdan2024collapse}, studies stability conditions and scaling \citep{bertrand2024stability,dohmatob2024demystified,dohmatob2024tale,dohmatob2024strong,seddik2024bad}, and curates or corrects such data \citep{feng2024beyond,gillman2024selfcorrecting}. \citet{schaeffer2025position} review the definitions in use. Repeated data have costs: duplicated sequences are memorised verbatim \citep{lee2022deduplicating} and a small repeated fraction can hurt a model beyond its share of tokens \citep{hernandez2022scaling}, while a few epochs over a whole corpus cost little \citep{muennighoff2023scaling}.

A related line documents the opposite failure, outputs that change between runs meant to be identical because batching changes floating-point reductions \citep{atil2024nondeterminism,he2025nondeterminism}. The hazard here is the converse of that failure: runs meant to differ are made identical. Samplers differ on this point. Hugging Face \texttt{generate} draws one \texttt{torch.multinomial} over the whole batch and keeps the rows of a batch independent \citep{wolf2020transformers}. SGLang \citep{zheng2024sglang} gives the parallel samples of one request a single seed when deterministic inference is enabled, and an open pull request derives a distinct seed for each \citep{sglang2025seedpr}. A reinforcement-learning codebase reports engine replicas that share one noise stream \citep{unirl2025seed}, and the vLLM reproducibility guide does not discuss the cross-request case \citep{vllm2025repro}. Simulation couples runs on purpose: common random numbers drive two systems with the same draws to reduce the variance of their comparison \citep{glasserman1992crn}.

Watermarking seeds a generator with a key and the preceding tokens and uses it to favour a keyed subset of the vocabulary \citep{kirchenbauer2023watermark} or as the sampling noise itself, so identical prompts yield identical outputs \citep{fu2024gumbelsoft}. The exponential-minimum and EXP watermarks are the exponential race of \S\ref{sec:mechanism} with keyed draws \citep{aaronson2022watermark,kuditipudi2024robust}. Models fine-tuned on watermarked text learn the watermark \citep{gu2024learnability,sander2024radioactive}. A shared seed plants an unkeyed watermark in a batch, and a self-training loop learns it.

\section{Setup}
\label{sec:setup}

\subsection{Models and the self-training loop}
\label{sec:loop}

We use eight public base checkpoints from a panel of thirteen of 1 to 4 billion parameters (Table~\ref{tab:spectrum}). Under the replayed seed of \S\ref{sec:mechanism} the self-loops of StableLM-2-1.6B, StableLM-3B, Minitron-4B and Granite-3.3-2B lose lexical diversity fastest in the panel. Those of Qwen3-1.7B, Qwen2.5-1.5B and SmolLM3-3B, chosen to add the Qwen and SmolLM families, lose it more slowly, and that of SmolLM2-1.7B slowest. A self-loop chain starts at generation 0 (g0) from the base model. In each generation the model continues 4{,}200 prompts, a quality filter keeps a pool of 2{,}100 continuations, and a fresh copy of the base weights is fully fine-tuned on the pool to give the next generation's model. Only text crosses generations. An ecosystem replaces the single model by $K$ members that write into one 2{,}100-row pool at fixed shares and are each fine-tuned on it. We run two with $K=3$, E3 with equal shares (SmolLM2-1.7B, StableLM-2-1.6B and StableLM-3B) and K3 with shares of 0.36, 0.36 and 0.28 (Qwen3-1.7B, SmolLM2-1.7B and Phi-2). A thirteen-member ecosystem of the whole panel, with Phi-2 at 0.28 and every other member at 0.06, runs to g5 under the replayed rule at three chain seeds, and Table~\ref{tab:spectrum} is ordered by each member's lexical diversity at g5.

Prompts are 8- to 16-word prefixes of WikiText-103 paragraphs \citep{merity2017pointer}. Generation uses vLLM 0.10.2 with temperature 1.0, top-p 0.95, repetition penalty 1.15, frequency penalty 0.3 and at most 128 new tokens. Fine-tuning runs one full-parameter epoch of AdamW \citep{loshchilov2019decoupled} at learning rate $2\times10^{-5}$ (Appendix~\ref{app:setup}).

We read every generation on the first 800 continuations of its pool. The main statistic is u4, the fraction of unique word 4-grams. Self-BLEU-4 \citep{zhu2018texygen}, GPT-2-large \citep{radford2019language} perplexity and the centroid drift of frozen DeBERTa-v3 embeddings \citep{he2023debertav3} from g0 ($1-\cos$) complement it. Two statistics count synchronisation: first-three sharing, the share of samples whose first three words match another's, and the document frequency of the top 4-gram.

\subsection{Seed handling and the four seed rules}
\label{sec:mechanism}

For a next-token distribution $p$, draw $q_v\sim\mathrm{Exp}(1)$ independently for every id $v$ and emit $\arg\max_v p_v/q_v$. The emitted id follows $p$ exactly, because $q_v/p_v$ is exponential with rate $p_v$ and the smallest of independent exponentials falls on $v$ with probability $p_v$ \citep{maddison2014astar,kool2019stochastic}. vLLM \citep{kwon2023vllm} samples this way. Requests without a seed are sampled in one batched call, whose rows are independent. A request that carries a seed gets its own \texttt{torch.Generator} seeded with it, which makes its output independent of the rest of the batch (Appendix~\ref{app:lottery}).

\texttt{LLM.generate} applies one \texttt{SamplingParams} object to every prompt it receives, and vLLM offsets the seed only between the $n$ parallel samples, or children, of one request. When the object carries a seed, every request builds a generator in the same state, draws the same vector $q(t)$ at its own step $t$, and advances by the same offset, so the draws depend only on the seed, the step and the id. Reconstructed from seed 42, they are bit-identical on every id below 32{,}000 across vocabulary widths from 32k to 256k, so checkpoints with different tokenizers share them as well (Appendix~\ref{app:lottery}).

Fix a step and an id $v$ whose draw at that step is $x$. In a request where $v$ has probability $p$, the other ids' draws let $v$ win with probability
$w_x(p)=\exp\!\big(-x(1-p)/p\big)$, and $\mathbb{E}_{x\sim\mathrm{Exp}(1)}\,w_x(p)=p$.
For a fixed model that does not depend on the current draws, each request is marginally an exact sample, although requests that share the seed are not jointly independent. Once the model has been fine-tuned on text written from the same draws, even this marginal exactness no longer holds conditional on the learned model. No test on one request of a fixed model can detect the coupling. Across requests, however, $x$ is shared. An id that draws a small $x$ at step $t$ wins at step $t$ far more often than its probability says, in every request that gives it moderate probability and whatever the prompt. The batch is a set of exact samples holding one lottery ticket per step, and the lottery writes the same ids at the same steps into unrelated continuations. We call such an (id, step) pair step-locked.

Each chain has a chain seed $s$, 42 unless stated, and four rules turn it into request seeds. Two of them share one seed across a batch. The replayed rule gives every request of every generation the seed $s$. The fresh rule gives every request of generation $g$ the seed $s+1000g$. Under the replayed rule generation $g{+}1$ meets the lottery its model was trained on, and under the fresh rule it meets a new one. We call the feedback that the replayed rule creates resonance: the sampling error of each generation acts as a fixed bias, and the next fine-tune reinforces it (\S\ref{sec:theory}). The other two rules give each request its own seed. The independent rule is the control and gives request $i$ of generation $g$ the seed $s\cdot10^6+g\cdot10^5+i$. The fixed per-request rule gives request $i$ the seed $s\cdot10^6+i$ in every generation, so each request replays draws that no other request shares. A coupling fraction $f$ mixes the replayed and the independent rule: a fixed random set of $\lceil fN\rceil$ of the $N=4{,}200$ requests receives $s$, the same set in every generation, and the rest receive independent seeds, so $f=1$ is the replayed rule and $f=0$ the independent rule.

The experiments ran on two GPU clusters, A and B, and Appendix~\ref{app:versions} lists which ran where. Rerunning a replayed self-loop with the same seed changes u4 at g3 by up to 0.017 (Table~\ref{tab:Eselfloops}), and we treat smaller differences as rerun noise. The same replayed self-loop differs by up to 0.07 at g3 between the clusters (Tables~\ref{tab:spectrum} and~\ref{tab:Gothers}), so each contrast below pairs chains from one cluster unless stated.

\section{Results}
\label{sec:results}

\subsection{Independent seeds prevent the fast loss of lexical diversity}
\label{sec:necessary}

\begin{figure}[t]
\centering
\figorbox{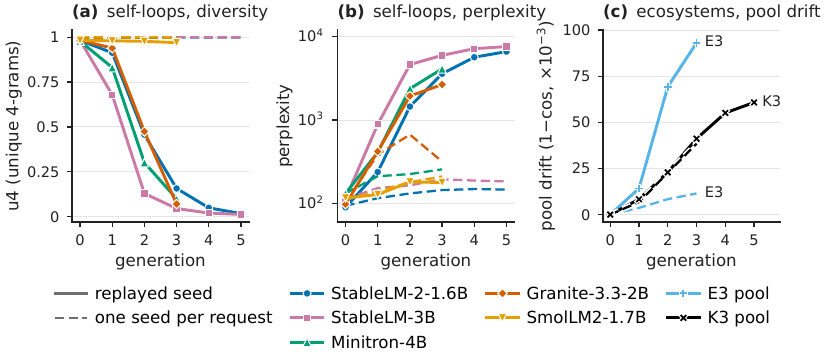}{\linewidth}{2.4in}
\caption{Replayed against independent seeds. (a) u4 and (b) GPT-2-large perplexity (log scale) of five self-loops, one colour and marker per checkpoint. Solid: the replayed rule. Dashed: the independent rule, in the colour of the same checkpoint. In (b) the dashed peak near 670 at g2 is Granite-3.3-2B turning incoherent (\S\ref{sec:remains}). (c) Drift of the pool centroid from g0 in the two $K=3$ ecosystems, with the same line styles. Per-generation values are in Tables~\ref{tab:Eselfloops} and~\ref{tab:Ek3}.}
\label{fig:necessary}
\end{figure}

In our self-loops the fast loss of lexical diversity requires the replayed seed. Rerunning them with independent seeds and nothing else changed keeps the diversity of all seven checkpoints that lose it under the replayed seed, from five families. Under the replayed seed their u4 at g3 lies between 0.045 and 0.38 (Figure~\ref{fig:necessary}a, Tables~\ref{tab:Eselfloops} and~\ref{tab:Gothers}), and under independent seeds on the same cluster it is at or above 0.98. At chain seed 43 the contrast holds for StableLM-2-1.6B, Minitron-4B and Granite-3.3-2B (Tables~\ref{tab:Eselfloops} and~\ref{tab:Gothers}). There the replayed seed takes Minitron-4B and Granite-3.3-2B below u4 of 0.07 at g3, and independent seeds keep both at or above 0.997. The contrast also holds for StableLM-2-1.6B on both clusters, and for both StableLM checkpoints through five generations. Without a shared stream no phrase is carried by many samples at once, and fine-tuning has no such phrase to amplify.

The same holds when collapsing checkpoints share a pool with one that does not. In E3, independent seeds keep every member at u4 of at least 0.999, and the pool centroid moves far less than under the replayed seed (Figure~\ref{fig:necessary}c, Table~\ref{tab:Ek3}), where the two StableLM members collapse and carry the pool along. In a mixed pool the seed acts through the members it breaks.

\subsection{The fingerprint belongs to the seed}
\label{sec:fingerprint}

A batch sampled under one seed carries step-locked phrases before any training, and these phrases belong to the seed. Of the g0 samples that StableLM-2-1.6B, StableLM-3B, Qwen3-1.7B, Qwen2.5-1.5B and SmolLM3-3B draw under one shared seed, 6\% to 9\% open with the same three words as another sample, against at most 0.3\% with a seed per request (Tables~\ref{tab:Gchains} and~\ref{tab:Gothers}). Every other checkpoint of the panel shares openings at least as often under the shared seed (Table~\ref{tab:spectrum}). Nothing is trained at g0, so these phrases are the lottery of \S\ref{sec:mechanism} in text.

The rebuilt lottery predicts the sampled tokens, even those the model did not rank first. Replaying the original g0 continuations of StableLM-2-1.6B and SmolLM2-1.7B through their base models, with the draws rebuilt from seed 42 and one backtracking step per derailed row, recovers 98\% of the tokens and at least 97\% of those that were not the most probable choice at their step, where draws from another seed, or the next step's draws, recover 4\% of the latter (Table~\ref{tab:Freplay}). Frequent ids that receive one of the smallest draws at a step are step-locked there almost always, and the share falls steeply as the draw grows (Table~\ref{tab:Fdose}). Because the draws depend on the seed, the step and the id and not on the checkpoint, checkpoints that share an id map, the tokenizer's assignment of strings to ids, lock the same ids at the same steps and write the same words, while checkpoints with different maps lock at the same steps but write different strings (Table~\ref{tab:Fcolock}). The phrases that later dominate a collapsed chain are born at a fixed step in the same way (Appendix~\ref{app:lottery}). The repeated phrases of a collapsed chain are properties of the noise before they are properties of the text.

\subsection{The loop resonates with a replayed realisation}
\label{sec:resonance}

\begin{figure}[t]
\centering
\figorbox{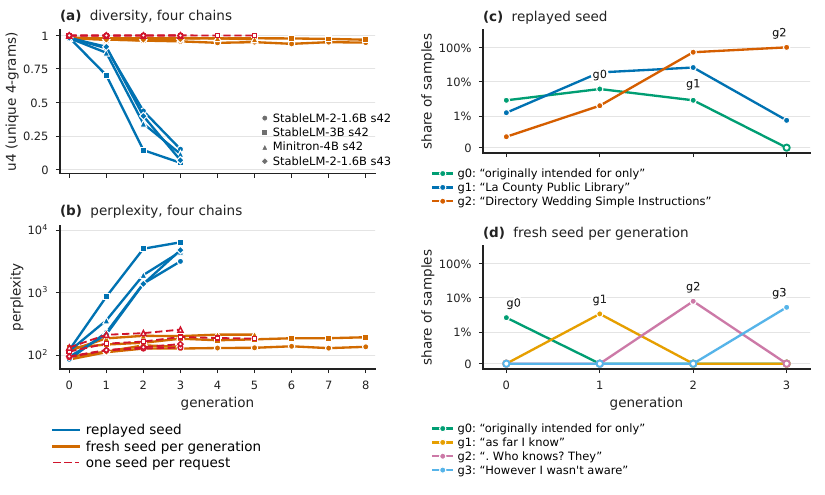}{\linewidth}{3.3in}
\caption{Replayed against fresh seeds on cluster B. (a) u4 and (b) GPT-2-large perplexity (log scale) of four chains, StableLM-2-1.6B at chain seeds 42 and 43, StableLM-3B and Minitron-4B, one marker per chain. Blue: the replayed rule. Orange: the fresh rule. Red dashed: one seed per request (cluster B for StableLM-2-1.6B at seed 42, cluster A otherwise). (c, d) The most frequent word 4-gram of each generation of the StableLM-2-1.6B chain (seed 42), traced across g0 to g3 under (c) the replayed seed and (d) a fresh seed per generation. Each line follows one phrase, tagged at the generation in which it is most frequent. Log scale, open markers at zero on the bottom tick. Tables~\ref{tab:trace} and~\ref{tab:dominant} give the plotted values and the top-ten lists.}
\label{fig:resonance}
\end{figure}

\begin{table}[t]
\centering
\caption{The four seed rules cross two properties of a seed, whether the batch shares it and whether it returns in every generation. Cells give u4 at g0 $\to$ g3 at chain seed 42 and the last column each chain's cluster, StableLM-2-1.6B first. Same-seed reruns differ in u4 at g3 by up to 0.017 (Table~\ref{tab:Eselfloops}).}
\label{tab:twobytwo}
\small
\begin{adjustbox}{max width=\linewidth}
\begin{tabular}{lccccc}
\toprule
seed rule & shared by the batch & returned every generation & StableLM-2-1.6B & StableLM-3B & cluster \\
\midrule
replayed & yes & yes & 0.981 $\to$ 0.152 & 0.983 $\to$ 0.051 & B \\
fresh & yes & no & 0.981 $\to$ 0.955 & 0.983 $\to$ 0.981 & B \\
fixed per-request & no & yes & 0.999 $\to$ 0.998 & 0.999 $\to$ 0.999 & B \\
independent & no & no & 0.999 $\to$ 0.999 & 0.999 $\to$ 1.000 & B, A \\
\bottomrule
\end{tabular}
\end{adjustbox}
\end{table}

The fast loss needs the draws to return. Under the fresh rule each generation carries step-locked phrases of its own (Figure~\ref{fig:resonance}d), which the next generation's draws do not favour. On StableLM-2-1.6B at two chain seeds, StableLM-3B and Minitron-4B, the replayed rule ends between 0.05 and 0.15 at g3, while the fresh rule keeps all four chains above 0.92, through g8 at seed 42 (Figure~\ref{fig:resonance}a,b, Table~\ref{tab:twobytwo}). At the shares one generation's draws write, duplicated text does not start the resonance.

Feedback needs the same realisation twice. Fine-tuning on generation $g$ raises the probability of the ids that won its lottery. Under the replayed rule, generation $g{+}1$ draws the same lottery, so those ids win again, by a larger margin and in more requests, and the next fine-tune raises them further. In each of the four replayed chains of Figure~\ref{fig:resonance} one phrase ends in 99\% to 100\% of samples at g3, and in the slower replayed chains of Qwen3-1.7B, Qwen2.5-1.5B and SmolLM3-3B in 79\% to 93\% (Tables~\ref{tab:Gchains} and~\ref{tab:Gothers}). Replay without sharing does not resonate: with a fixed seed per request, StableLM-2-1.6B and StableLM-3B keep u4 at or above 0.998 at g3 (Table~\ref{tab:twobytwo}), since an id's draws differ across requests and its share stays near its probability. The loop needs a draw both shared and returned. Repeated prompts alone do not start it, and at temperature 0.7 the replayed chains track the independent ones through six generations (Appendix~\ref{sec:nontriggers}).

A new shared seed per generation slows the resonance, and a phrase can still grow under it. The fresh seed ends the replay but keeps the sharing, so each generation still writes step-locked phrases into many samples, and the next fine-tune raises them. Of three fresh chains run to g8, StableLM-2-1.6B keeps its top phrase under 8\% of samples, while ``British Army Chief General'' enters StableLM-3B at g4 and grows to 15\% at g8, and one phrase of Minitron-4B jumps to 30\% at g8, with more than half of that generation's samples sharing their first three words with another (Tables~\ref{tab:Gchains}, \ref{tab:Gothers} and~\ref{tab:freshlong}). One reading is that a phrase grows when a later draw happens to favour ids an earlier fine-tune raised (\S\ref{sec:theory}). Growth is then left to chance, and only a seed per request removes the push.

The template a replayed chain converges to comes from the draws. Table~\ref{tab:qualitative} in Appendix~\ref{app:phrases} traces the replayed StableLM-2-1.6B template to its origin. Its core is a string of capitalised words that occurs in none of the prompts and in no g0 sample drawn with independent seeds. The seed-42 lottery writes it into two g0 continuations, about 75 words in, and the phrase keeps that slot in every later generation while the spread of its position narrows. Only 12\% of g3 samples share even their first three words with another (Table~\ref{tab:Gchains}), so most samples reach the template through different openings and meet it where the lottery placed it. A replayed loop grows a string that one realisation of the noise wrote.

\subsection{Collapse grows with the coupled share and spreads late}
\label{sec:dose}

\begin{figure}[t]
\centering
\figorbox{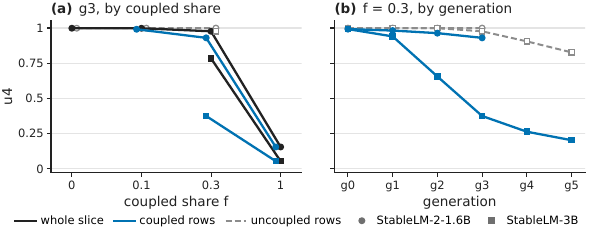}{0.72\linewidth}{1.55in}
\caption{Partial coupling of StableLM-2-1.6B (circles) and StableLM-3B (squares, $f=0.3$ and $f=1$ only). (a) u4 at g3 against the coupled share $f$ for the whole slice (black), the coupled rows (blue) and the uncoupled rows (grey dashed). (b) u4 of the coupled and uncoupled rows by generation at $f=0.3$.}
\label{fig:dose}
\end{figure}

Inside the requests coupled to the replayed seed, diversity falls faster the larger the coupled share. With three tenths of the batch coupled, StableLM-3B's coupled rows already reach u4 of 0.37 at g3, while StableLM-2-1.6B's fall more slowly (blue in Figure~\ref{fig:dose}a, Tables~\ref{tab:Gchains} and~\ref{tab:Gothers}). The whole slice of StableLM-2-1.6B, which mixes coupled and uncoupled rows, changes little up to three tenths and collapses only under full coupling (black in Figure~\ref{fig:dose}a). This is consistent with a fine-tune that raises a phrase in proportion to its share of the pool, so that the coupled share sets the gain of the loop.

The collapse starts inside the coupled requests and reaches the others only after it has made its phrases common in the pool, although both come from the same fine-tuned model in the same engine call. At g3 of the StableLM-2-1.6B chain at $f=0.3$, the dominant phrase ``La County Public Access'' is in 43\% of the coupled rows and in none of the uncoupled rows. Independent draws reproduce a rare phrase far below its share of the pool while the replayed draws amplify it many times over, and only once the phrase holds about 15\% of the pool do independent draws return it at half to three quarters of that share (Table~\ref{tab:split}). In StableLM-3B at the same coupling the uncoupled rows lose diversity about two generations after the coupled rows (Table~\ref{tab:Gothers}). The resonance makes a phrase common, and ordinary self-training on the duplicated text then spreads it, more slowly, to the requests the replay never reached.

\subsection{The weights carry the phrases, and the shared draw makes them universal}
\label{sec:xdec}

\begin{table}[t]
\centering
\caption{Generation-3 weights of the StableLM-2-1.6B replayed and independent chains (chain seed 42), each decoded on the generation-3 prompts under its own seed rule and under the other, with the base weights under both rules. u4 and the document frequency of the most common 4-gram over the 800 continuations, the share of continuations containing the phrase that fills every sample of the replayed chain, and perplexity under GPT-2-large.}
\label{tab:xdec}
\small
\begin{tabular}{llrrrr}
\toprule
weights & decoded with & u4 & top df & phrase & PPL \\
\midrule
replayed chain & shared seed 42 & 0.153 & 1.000 & 100\% & 3193 \\
replayed chain & one seed per request & 0.822 & 0.456 & 46\% & 842 \\
\addlinespace[2pt]
independent chain & one seed per request & 0.998 & 0.009 & 0\% & 128 \\
independent chain & shared seed 42 & 0.969 & 0.036 & 0\% & 133 \\
\addlinespace[2pt]
base (generation 0) & shared seed 42 & 0.981 & 0.029 & 0\% & 86 \\
base (generation 0) & one seed per request & 0.999 & 0.006 & 0\% & 93 \\
\bottomrule
\end{tabular}
\end{table}

How much of the loss sits in the weights and how much in the sampler that meets them can be read by decoding one checkpoint under the other rule. We retrained the generation-3 weights of the StableLM-2-1.6B replayed chain and of its independent-seed chain from the base, each on its own generation-2 pool, and decoded each on the generation-3 prompts under its own seed rule and under the other (Table~\ref{tab:xdec}). Under their own rules the retrained weights reproduce the chains, u4 0.153 against 0.152 and 0.998 against 0.998. Decoded with one seed per request, the replayed weights recover most of their diversity, u4 0.82, yet the phrase that fills every sample under the shared seed still opens 46\% of them, a second phrase 39\%, and perplexity under GPT-2-large stays at 842 against 128 for the independent weights. The independent weights decoded under the shared seed lose 0.03 of u4 and share their most common 4-gram in 3.6\% of samples, about what the base loses under the same seed. Three generations of the loop have written the favoured phrases into the weights at probabilities that no longer need the shared draw to appear in nearly half the samples. The shared draw turns that half into all, and its return in every generation is what raised the probabilities.

\subsection{What remains without the resonance}
\label{sec:remains}

For five of the eight checkpoints, what survives independent seeds is slow drift. The self-trained models still move: in every independent-seed self-loop of these five, perplexity at g3 lies between a third above and about twice its g0 value and continuations lengthen (Figure~\ref{fig:necessary}b, Tables~\ref{tab:Eselfloops}, \ref{tab:Gchains} and~\ref{tab:Gothers}). In K3, whose members do not collapse into a template within three generations even under the replayed seed, the pool centroid drifts as far under independent seeds as under the replayed seed while the loss of pool u4 disappears (Table~\ref{tab:Ek3}). This slow drift needs no correlated noise, nor, for StableLM-2-1.6B, any memory in the weights: when each generation fine-tunes the previous generation's model instead of the base, u4 and perplexity match the restarted chain at g3 and stay level to g5 (Table~\ref{tab:Gchains}). It is consistent with the gradual accumulation of per-generation error that theory describes, as in the gradual loss of lexical diversity reported by \citet{guo2024curious}, and the replayed stream adds a fast, phrase-level channel.

The other three checkpoints lose coherence within three generations without the seed. Granite-3.3-2B keeps u4 near 1, yet its text turns into word salad at g2 and into bracket and punctuation loops at g3, when 93\% of its samples contain a long run of symbols (Appendix~\ref{app:setup}). Qwen3-1.7B and Qwen2.5-1.5B follow more slowly, into word lists and switches of language, and their perplexity at g3 is 2.6 and 3.5 times its g0 value (Table~\ref{tab:Gothers}). Perplexity and embedding drift register the change (Table~\ref{tab:Eselfloops}), while every sample still differs from every other, so u4 scores these chains as diverse. In these checkpoints a loss of coherence that needs no shared seed runs alongside the resonance.

\subsection{The same pattern in a published pipeline}
\label{sec:wild}

One peer-reviewed recursive-training pipeline broadcasts one seeded configuration to a batch, and its released data hold copies of the size one such batch writes. The sampling script of \citet{kazdan2024collapse} sends 64 identical empty prompts per call through \texttt{llm.generate} with one \texttt{SamplingParams(n=64, seed=k)}, where $k$ restarts from zero in every run, so each iteration meets the same seeds again. Executed under the pinned vLLM 0.5.4, this call returns 64 distinct continuations, one per child, each copied into every request (Table~\ref{tab:wild} in Appendix~\ref{app:published}), and in the released Gemma-2-27B data the dominant copy multiplicities are consistent with that duplication: at iteration 2 with SFT seed 0, three quarters of the rows are texts that occur 54 to 60 times, close to the 64 copies per call, and the next round of training takes them as data. With distinct prompts, as in our pipeline, a shared seed writes shared phrases instead of copies, and resonance makes them grow (\S\ref{sec:resonance}). The audit shows that the coupling pattern occurs in a published pipeline, not how much of that work's reported behaviour is due to seed replay.

\section{A stylised reading}
\label{sec:theory}

Stability analyses of recursive training model each generation as a fresh sample from the current model \citep{bertrand2024stability,gerstgrasser2024collapse,dohmatob2024tale}, and a replayed stream breaks that assumption. Take one id at one step with the same probability $p_g$ in every request at generation $g$, and let $x$ be its draw. A shared $x$ writes the id into about a share $w_x(p_g)$ of the samples (\S\ref{sec:mechanism}), and if a fine-tune closes a fraction $\eta\in(0,1]$ of the gap to that share,
\begin{equation}
p_{g+1}\approx p_g+\eta\,\big[w_x(p_g)-p_g\big],\qquad
w_x(p)>p \iff x<h(p)=\frac{p\ln(1/p)}{1-p}.
\label{eq:loop}
\end{equation}
where $h(p)$ is the largest draw at which an id of probability $p$ still gains share. Eq.~(\ref{eq:loop}) is a phenomenological description of the response across generations, not the update of the fine-tune, which restarts from the base weights in every generation. The share $w_x(p)$ averages over the draws of the competing ids, so it is not the batch frequency under one full shared draw vector, and the fixed points and variances below hold within this scalar model. With a fresh $x$ in every generation, or a different $x$ in every request, the bracket has mean zero, because $\mathbb{E}_x w_x(p)=p$, and the sampling error only diffuses $p$. A replayed $x$ keeps the sign of the bracket from one generation to the next. In the model, since $h$ rises from 0 to 1 on $(0,1)$, a draw $x\ge1$ favours no id at any $p$, and for $x<1$ the fixed points $p=0$ and $p=1$ attract and $h^{-1}(x)$ repels. In the data, the ids the shared lottery favours have the smallest draws, on the growing branch $x<h(p)$ (Table~\ref{tab:Fdose}), and under the replayed seed the top phrase of StableLM-2-1.6B climbs to every sample, toward $p=1$, while under a fresh seed its top phrases do not persist (Figure~\ref{fig:resonance}c,d). In the model a fresh shared $x$ removes the bias but keeps the variance of a shared draw, $\eta^2p(1-p)^2/(2-p)$ per generation, about a thousand times the $\eta^2p(1-p)/2{,}100$ of a pool of 2{,}100 independent samples when $p$ is small. The largest share among many ids still drifts upward, as the maximum of unbiased random walks does, and faster the wider their spread, which fits the fresh StableLM-3B phrase that grows from g4 (\S\ref{sec:resonance}). In Minitron-4B, 58\% of the g8 samples share their first three words with another, against at most 12\% before (Table~\ref{tab:Gothers}). Its jump is written by one generation's draws.

The model orders the seed rules and says nothing about which checkpoint or which phrase wins. With one id and a fixed $\eta$ it cannot say which of several step-locked phrases takes over, and a common $\eta$ does not rank the checkpoints: in the thirteen-member ecosystem, the checkpoints whose g0 samples share the most openings under the shared seed keep the most diversity, with a rank correlation of $+0.65$ (Appendix~\ref{app:spectrum}), where Eq.~(\ref{eq:loop}) predicts the reverse. Within the model the ordering must come from how far each checkpoint's fine-tune moves toward the shared draws, a response the model takes as given.

\section{Discussion and conclusion}
\label{sec:discussion}

Any loss of lexical diversity within a few generations is a candidate for resonance, and three checks tell whether the sampler can be its cause: whether one sampling object with a seed is broadcast to a batch, whether that seed changes between generations, and whether samples of different prompts share their first words far more often than under independent sampling. A fresh shared seed per generation keeps the fingerprint and slows the resonance, yet in two of three long chains a phrase still reaches 15\% and 30\% of samples under it (\S\ref{sec:resonance}). The fix is a seed per request, derived from the chain seed, the generation and the request index.

\paragraph{Limitations.} Each cell of the self-loop and ecosystem experiments is one chain. A second chain seed, 43, repeats the contrast of replayed and independent seeds for StableLM-2-1.6B, Minitron-4B and Granite-3.3-2B and the fresh rule for StableLM-2-1.6B, all to g3. Every other chain uses chain seed 42 alone, independent seeds were followed to g5 only on the two StableLM checkpoints, the fresh rule past g3 only on these two and Minitron-4B, temperature 0.7 only on these two, and the decoding of one checkpoint under the other seed rule covers StableLM-2-1.6B at one chain seed and one generation. The self-loops use eight checkpoints of 1 to 4 billion parameters, WikiText-103 prompts and one fine-tuning recipe whose training seed equals the chain seed, and whether the incoherence of Granite-3.3-2B and the two Qwen checkpoints depends on recipe or prompts is untested. Every chain ran on vLLM 0.10.2 with explicit seeds, the published call pattern was executed with a stand-in model, and the recount of the published data cannot separate the copies a shared seed writes from those a model would write under any seed. Beyond the pipeline of \S\ref{sec:wild}, we did not check which published fast collapses used a replayed seed.

Recursive training has been read as a property of models and data. Under a replayed seed it is also a property of the sampler, and the remedy is the one armies adopted on bridges: break step.

\clearpage
\subsubsection*{Ethics statement}
This work trains and evaluates publicly released open-weight checkpoints on public corpora and involves no human subjects, personal data or annotators. Section~\ref{sec:wild} names a public repository and recounts its public data to show that the seeding pattern occurs in practice. The recount describes the released data and does not re-examine the conclusions of the work that produced them.

\subsubsection*{Reproducibility statement}
Appendix~\ref{app:setup} gives the checkpoints, prompts, filter and recipes, Appendix~\ref{app:spectrum} the Hugging Face ids, Appendix~\ref{app:versions} the software and hardware, and Appendix~\ref{app:repro} the seed code of each rule. The per-generation tables of Appendix~\ref{app:tables} hold every value plotted in Figures~\ref{fig:necessary}, \ref{fig:resonance}a,b and~\ref{fig:dose}, and Table~\ref{tab:trace} those of Figure~\ref{fig:resonance}c,d. Code, prompt files and measurement files will be released.

\subsubsection*{AI use statement}
In this work, we used generative AI tools to implement standard components of the experimental pipeline, to assist with data analysis and figure drawing, and to draft and edit parts of the paper. The recursively generated text that this paper studies is the object of study itself: it is produced by the open-weight checkpoints under examination, following the procedure of \S\ref{sec:setup} and Appendix~\ref{app:setup}. We have reviewed all AI-assisted work: the analysis scripts were rerun by the authors on the stored measurements, and all drafted and edited text was read and approved by the authors. We take responsibility for the final content of this work, including text, claims or artifacts produced with the aid of generative AI.

\bibliography{references}
\bibliographystyle{iclr2027_conference}

\appendix

\section{Setup details}
\label{app:setup}

\paragraph{Checkpoints.}
The self-loops use StableLM-2-1.6B \citep{bellagente2024stable}, StableLM-3B \citep{tow2023stablelm3b}, Minitron-4B \citep{muralidharan2024compact}, Granite-3.3-2B \citep{granite2025granite33}, Qwen3-1.7B \citep{yang2025qwen3}, Qwen2.5-1.5B \citep{qwen2024qwen25}, SmolLM3-3B \citep{bakouch2025smollm3} and SmolLM2-1.7B \citep{benallal2025smollm2}, and K3 adds Phi-2 \citep{javaheripi2023phi2}. The rest of the panel is OLMo-2-1B \citep{olmo2024olmo2}, Falcon3-1B \citep{falcon2024falcon3}, Tucano-2B4 \citep{correa2025tucano} and Kumru-2B \citep{turker2025kumru}. Table~\ref{tab:spectrum} in Appendix~\ref{app:spectrum} gives the Hugging Face id of each.

\paragraph{Generation.}
Prompts are 8- to 16-word neutral prefixes of WikiText-103 paragraphs, frozen into one file per generation and chain seed and shared by all models of a generation. A self-loop generation writes 4{,}200 continuations. A label-blind filter keeps continuations with at least 20 words, a distinct-bigram ratio of at least 0.45 and no single word above 20\% of the words, and the first 2{,}100 kept continuations form the pool. The filter judges each continuation on its own, so a phrase repeated across samples passes it, and in every generation of the cluster-B chains, collapsed ones included, it kept at least 2{,}100 continuations. The first 800 rows of the pool form the measured slice. An ecosystem member writes 800 continuations per generation, and the pool draws from the members at their shares with largest-remainder rounding to 2{,}100 rows. The sampler settings are those of \S\ref{sec:setup}. Coupled and uncoupled flags are assigned per raw request before the filter, so the pool and the slice inherit them row by row.

\paragraph{Fine-tuning.}
Every generation starts from the clean base weights, except in the chained-weights chain of Table~\ref{tab:Gchains}, where the model of each generation from g2 on is fine-tuned from the model of the generation before. Full-parameter fine-tuning runs one epoch with AdamW ($\beta_1=0.9$, $\beta_2=0.999$, $\epsilon=10^{-8}$), learning rate $2\times10^{-5}$ with 40 warm-up steps and linear decay, weight decay 0.01, bf16, gradient checkpointing and maximum length 768. The per-device batch is 8 with 2 accumulation steps, an effective batch of 16, and checkpoints of 3B parameters and more halve the batch and double the accumulation. Each row is the prompt followed by its continuation, with no packing and no masking of the prompt. The training seed equals the chain seed in every generation and under every seed rule, so the chains that keep their diversity under independent or fixed per-request seeds replay it too, and a replayed training seed alone does not produce the fast loss.

\paragraph{Measurement.}
u4 is the number of distinct word 4-grams divided by the number of word 4-grams in the 800-row slice. Self-BLEU-4 uses 80 texts. Perplexity is per GPT-2 token under GPT-2-large on 200 texts. Embedding drift is $1-\cos$ between the slice centroid at generation $g$ and at g0 under the frozen encoder \texttt{MoritzLaurer/DeBERTa-v3-base-mnli-fever-anli}. The symbol-run share of \S\ref{sec:remains} is the share of samples containing a run of six or more characters that are neither alphanumeric nor whitespace. Under independent seeds it is 1\%, 8\%, 34\% and 93\% at g0 to g3 for Granite-3.3-2B (2\%, 6\%, 25\% and 78\% at chain seed 43), 4\%, 8\%, 17\% and 37\% for Qwen3-1.7B and 2\%, 4\%, 11\% and 19\% for Qwen2.5-1.5B, and it lies between 1\% and 4\% in every generation of the other independent-seed self-loops.

\section{The checkpoint panel}
\label{app:spectrum}

\begin{table}[h]
\centering
\caption{The checkpoint panel: thirteen base checkpoints of 1 to 4 billion parameters with their Hugging Face ids, ordered by the ecosystem endpoint under the replayed seed. The main text uses the four checkpoints with the lowest replayed self-loop u4 at g3, three from the middle of the replayed self-loop column that add the Qwen and SmolLM families to those four (Qwen2.5-1.5B, Qwen3-1.7B and SmolLM3-3B), and SmolLM2-1.7B, which has the highest. Eco g5: u4 at generation 5 of the thirteen-member ecosystem of \S\ref{sec:loop}, mean of three chain seeds. Self-loop g3: u4 at generation 3 of the seed-42 self-loop under the replayed rule, and under the independent rule where that chain was run. All values come from cluster A except those marked $^{b}$, from cluster B, where the replayed chains of SmolLM3-3B, Qwen2.5-1.5B and Qwen3-1.7B end at 0.377, 0.280 and 0.285 (Table~\ref{tab:Gothers}). First-three sharing: share of generation-0 samples whose first three words match another sample of the same stream, in the seed-42 ecosystem and in the self-loop.}
\label{tab:spectrum}
\small
\setlength{\tabcolsep}{3.5pt}
\begin{adjustbox}{max width=\linewidth}
\begin{tabular}{lllccccc}
\toprule
 & & & eco g5 & \multicolumn{2}{c}{self-loop g3 u4} & \multicolumn{2}{c}{first-three sharing, g0} \\
\cmidrule(lr){5-6}\cmidrule(lr){7-8}
name & Hugging Face id & size & u4 & replayed & independent & ecosystem & self-loop \\
\midrule
StableLM-2-1.6B & \texttt{stabilityai/stablelm-2-1\_6b} & 1.6B & 0.187 & 0.157 & 0.999 & 0.088 & 0.091 \\
SmolLM3-3B & \texttt{HuggingFaceTB/SmolLM3-3B-Base} & 3B & 0.227 & 0.329 & 0.999$^{b}$ & 0.088 & 0.081 \\
StableLM-3B & \texttt{stabilityai/stablelm-3b-4e1t} & 3B & 0.229 & 0.045 & 1.000 & 0.090 & 0.093 \\
Minitron-4B & \texttt{nvidia/Minitron-4B-Base} & 4B & 0.340 & 0.099 & 0.999 & 0.121 & 0.121 \\
Granite-3.3-2B & \texttt{ibm-granite/granite-3.3-2b-base} & 2.5B & 0.405 & 0.070 & 0.996 & 0.092 & 0.101 \\
Qwen2.5-1.5B & \texttt{Qwen/Qwen2.5-1.5B} & 1.5B & 0.498 & 0.214 & 0.992$^{b}$ & 0.064 & 0.068 \\
OLMo-2-1B & \texttt{allenai/OLMo-2-0425-1B} & 1B & 0.509 & 0.728 & -- & 0.091 & 0.093 \\
Phi-2 & \texttt{microsoft/phi-2} & 2.7B & 0.572 & 0.923 & -- & 0.205 & 0.204 \\
Qwen3-1.7B & \texttt{Qwen/Qwen3-1.7B-Base} & 1.7B & 0.576 & 0.272 & 0.981$^{b}$ & 0.088 & 0.094 \\
Tucano-2B4 & \texttt{TucanoBR/Tucano-2b4} & 2.4B & 0.713 & 0.300 & -- & 0.163 & 0.155 \\
Falcon3-1B & \texttt{tiiuae/Falcon3-1B-Base} & 1B & 0.745 & 0.532 & -- & 0.149 & 0.156 \\
Kumru-2B & \texttt{vngrs-ai/Kumru-2B-Base} & 2B & 0.867 & 0.541 & -- & 0.255 & 0.260 \\
SmolLM2-1.7B & \texttt{HuggingFaceTB/SmolLM2-1.7B} & 1.7B & 0.940 & 0.969 & 0.999 & 0.164 & 0.168 \\
\bottomrule
\end{tabular}
\end{adjustbox}
\end{table}

The eight checkpoints of the self-loops come from the panel of Table~\ref{tab:spectrum}, and the panel shows that susceptibility to the shared stream varies widely across checkpoints. The ecosystem column comes from the thirteen-member ecosystem of \S\ref{sec:loop}, run at chain seeds 42, 123 and 456. Under the replayed seed the thirteen checkpoints end their ecosystem chains at very different u4. For the seven checkpoints that collapse under the replayed seed and were rerun with independent seeds, this ordering reflects how fast their fine-tunes amplify the shared stream. A larger fingerprint goes with a slower collapse. The six checkpoints with the highest ecosystem endpoint are more synchronised at g0 than the six with the lowest, with mean first-three sharing of 0.170 against 0.090 in the seed-42 ecosystem, and first-three sharing at g0 has a rank correlation of $+0.65$ with the g5 endpoint across the thirteen ($p=0.02$, two-sided permutation test), so the checkpoints that the shared seed synchronises most keep the most diversity. With a common $\eta$, Eq.~(\ref{eq:loop}) would predict the reverse, so within the model the ordering must come from $\eta$, how far each checkpoint's fine-tune moves toward the shared draws (\S\ref{sec:theory}). On cluster B, StableLM-3B and StableLM-2-1.6B share openings equally often at g0 under the replayed seed, 9\% of samples each, yet StableLM-3B reaches u4 0.70 after one generation where StableLM-2-1.6B is at 0.90 (Tables~\ref{tab:Gchains} and~\ref{tab:Gothers}). Which property of a checkpoint sets its response is open.

\section{Software and hardware}
\label{app:versions}

\begin{table}[h]
\centering
\caption{Software and hardware of the two clusters. Both run the V1 engine of vLLM 0.10.2.}
\label{tab:versions}
\small
\begin{tabular}{p{0.15\linewidth}p{0.37\linewidth}p{0.37\linewidth}}
\toprule
 & cluster A (NVIDIA RTX 4090) & cluster B (NVIDIA L40, 46 GB) \\
\midrule
generation & vLLM 0.10.2, torch 2.8.0+cu128, transformers 4.56.2, Python 3.9, numpy 2.0.2 & vLLM 0.10.2, torch 2.8.0+cu128, xformers 0.0.32.post1, transformers 4.56.2, tokenizers 0.22.1, huggingface-hub 0.35.3, Python 3.10, numpy 2.2.6, Flash Attention backend \\
training & torch 2.4.0+cu121, transformers 5.0.0, accelerate 1.13.0, datasets 5.0.1, Python 3.10.12 & torch 2.4.0+cu121, transformers 5.0.0, accelerate 1.13.0, datasets 5.0.1, tokenizers 0.22.2, huggingface\_hub 1.17.0, Python 3.10 \\
experiments & independent seeds, ecosystems, lottery reconstruction, published call pattern & coupling fraction, replayed against fresh seeds, chain seed 43, fixed per-request seeds, prompt repetition and temperature, independent seeds for three further checkpoints, chained weights \\
\bottomrule
\end{tabular}
\end{table}

The published call pattern (Appendix~\ref{app:published}) ran on cluster A in a separate environment with vLLM 0.5.4 (V0 engine), torch 2.4.0+cu121 and transformers 4.44.2, and in the cluster-A environment of vLLM 0.10.2.

\section{Per-generation tables}
\label{app:tables}

Tables~\ref{tab:Eselfloops} and~\ref{tab:Ek3} give the cluster-A chains of \S\ref{sec:necessary}. Table~\ref{tab:Gchains} gives the cluster-B chains of StableLM-2-1.6B and Table~\ref{tab:Gothers} those of the other checkpoints. Rerunning three replayed self-loops with the same seed reproduces u4 at g3 to within 0.017, although vLLM reruns are not bit-identical (Appendix~\ref{app:lottery}).

\begin{table}[h]
\centering
\caption{Self-loops under the independent rule against the replayed rule (cluster A). Top: u4 of the 800-sample slice by generation and DeBERTa centroid drift from g0 at g3 ($\times10^{-3}$). Bottom: GPT-2-large perplexity by generation. ``Replayed, rerun'' is a second run of the replayed chain above it with the same seed. The suffixes s42 and s43 mark chain seeds 42 and 43. Dashes mark generations that were not run.}
\label{tab:Eselfloops}
\small
\setlength{\tabcolsep}{3pt}
\begin{adjustbox}{max width=\linewidth}
\begin{tabular}{llccccccr}
\toprule
chain & seed rule & g0 & g1 & g2 & g3 & g4 & g5 & drift g3 ($\times10^{-3}$) \\
\midrule
\multicolumn{9}{l}{\emph{u4}} \\
SmolLM2-1.7B s42 & independent & 0.999 & 0.999 & 0.999 & 0.999 & -- & -- & 23.8 \\
SmolLM2-1.7B s42 & replayed & 0.982 & 0.980 & 0.977 & 0.969 & -- & -- & 18.3 \\
SmolLM2-1.7B s42 & replayed, rerun & 0.983 & 0.979 & 0.977 & 0.973 & -- & -- & 23.9 \\
\addlinespace
StableLM-2-1.6B s42 & independent & 0.999 & 0.998 & 0.999 & 0.999 & 0.999 & 0.999 & 2.5 \\
StableLM-2-1.6B s42 & replayed & 0.982 & 0.913 & 0.456 & 0.157 & 0.049 & 0.017 & 181.5 \\
StableLM-2-1.6B s42 & replayed, rerun & 0.982 & 0.916 & 0.428 & 0.140 & -- & -- & 171.6 \\
\addlinespace
StableLM-3B s42 & independent & 0.999 & 0.999 & 0.999 & 1.000 & 0.999 & 0.999 & 3.8 \\
StableLM-3B s42 & replayed & 0.983 & 0.679 & 0.129 & 0.045 & 0.020 & 0.012 & 188.6 \\
StableLM-3B s42 & replayed, rerun & 0.983 & 0.680 & 0.134 & 0.046 & -- & -- & 206.2 \\
Granite-3.3-2B s42 & independent & 0.998 & 0.999 & 0.999 & 0.996 & -- & -- & 178.8 \\
Granite-3.3-2B s42 & replayed & 0.983 & 0.939 & 0.474 & 0.070 & -- & -- & 171.5 \\
Minitron-4B s42 & independent & 0.998 & 0.998 & 0.999 & 0.999 & -- & -- & 5.9 \\
Minitron-4B s42 & replayed & 0.978 & 0.831 & 0.301 & 0.099 & -- & -- & 153.8 \\
StableLM-2-1.6B s43 & independent & 0.999 & 0.999 & 0.999 & 0.999 & -- & -- & 2.4 \\
StableLM-2-1.6B s43 & replayed & 0.980 & 0.918 & 0.389 & 0.080 & -- & -- & 196.6 \\
\midrule
\multicolumn{9}{l}{\emph{GPT-2-large perplexity}} \\
SmolLM2-1.7B s42 & independent & 111 & 140 & 180 & 211 & -- & -- & \\
SmolLM2-1.7B s42 & replayed & 117 & 128 & 181 & 178 & -- & -- & \\
SmolLM2-1.7B s42 & replayed, rerun & 114 & 132 & 178 & 183 & -- & -- & \\
\addlinespace
StableLM-2-1.6B s42 & independent & 94 & 116 & 132 & 145 & 149 & 147 & \\
StableLM-2-1.6B s42 & replayed & 90 & 239 & 1{,}450 & 3{,}588 & 5{,}661 & 6{,}607 & \\
StableLM-2-1.6B s42 & replayed, rerun & 87 & 228 & 1{,}442 & 4{,}009 & -- & -- & \\
\addlinespace
StableLM-3B s42 & independent & 114 & 153 & 165 & 196 & 188 & 184 & \\
StableLM-3B s42 & replayed & 115 & 888 & 4{,}624 & 5{,}956 & 7{,}169 & 7{,}597 & \\
StableLM-3B s42 & replayed, rerun & 115 & 884 & 4{,}745 & 6{,}255 & -- & -- & \\
Granite-3.3-2B s42 & independent & 104 & 386 & 669 & 319 & -- & -- & \\
Granite-3.3-2B s42 & replayed & 98 & 420 & 1{,}940 & 2{,}664 & -- & -- & \\
Minitron-4B s42 & independent & 133 & 212 & 225 & 257 & -- & -- & \\
Minitron-4B s42 & replayed & 132 & 412 & 2{,}377 & 4{,}102 & -- & -- & \\
StableLM-2-1.6B s43 & independent & 96 & 121 & 134 & 150 & -- & -- & \\
StableLM-2-1.6B s43 & replayed & 91 & 213 & 1{,}357 & 3{,}878 & -- & -- & \\
\bottomrule
\end{tabular}
\end{adjustbox}
\end{table}

\begin{table}[h]
\centering
\caption{K$=$3 ecosystems under the independent rule (seeds distinct across prompts, generations and models) against the replayed rule. Drift is the DeBERTa centroid distance from g0 ($1-\cos$, $\times10^{-3}$).}
\label{tab:Ek3}
\small
\begin{adjustbox}{max width=\linewidth}
\begin{tabular}{lll}
\toprule
quantity & independent & replayed \\
\midrule
\multicolumn{3}{l}{\emph{E3 (SmolLM2-1.7B, StableLM-2-1.6B, StableLM-3B at one third each)}} \\
pool drift ($\times10^{-3}$), g0 to g3 & 0.0 / 3.6 / 8.4 / 11.5 & 0.0 / 14.2 / 69.3 / 93.0 \\
pool u4, g0 to g3 & 0.998 / 0.999 / 0.999 / 0.999 & 0.983 / 0.919 / 0.618 / 0.451 \\
member u4 at g3 (order as listed) & 0.999, 1.000, 1.000 & 0.955, 0.237, 0.137 \\
member drift at g3 (order as listed) & 13.1, 20.1, 8.8 & 15.8, 150.8, 177.3 \\
member PPL g0 $\to$ g3 (order as listed) & 121$\to$206, 96$\to$293, 128$\to$258 & 119$\to$240, 92$\to$2{,}980, 125$\to$2{,}856 \\
\addlinespace
\multicolumn{3}{l}{\emph{K3 (Qwen3-1.7B 0.36, SmolLM2-1.7B 0.36, Phi-2 0.28)}} \\
pool drift ($\times10^{-3}$), g0 to g3 & 0.0 / 6.8 / 22.6 / 38.4 & 0.0 / 8.3 / 22.9 / 41.1 \\
pool u4, g0 to g3 & 0.985 / 0.998 / 0.998 / 0.996 & 0.972 / 0.954 / 0.904 / 0.819 \\
pool drift ($\times10^{-3}$), g4 / g5 & -- & 55.1 / 60.8 \\
pool u4, g4 / g5 & -- & 0.734 / 0.699 \\
member u4 at g3 (order as listed) & 0.991, 0.999, 1.000 & 0.626, 0.954, 0.874 \\
member drift at g3 (order as listed) & 64.5, 21.4, 52.8 & 58.3, 22.9, 69.6 \\
\bottomrule
\end{tabular}
\end{adjustbox}
\end{table}

\begin{table}[p]
\centering
\caption{The cluster-B chains of StableLM-2-1.6B (800-sample slice, chain seed 42 unless marked). $f$ is the coupling fraction. Coupled and uncoupled columns give u4 within the rows that did and did not receive the batch's shared seed. PPL is GPT-2-large perplexity per token on 200 texts, words the mean continuation length, first-3 the share of samples whose first three words match another sample, top df the document frequency of the most frequent word 4-gram. In the chained-weights chain each generation from g2 on fine-tunes the previous generation's model instead of the base. In the prompts $\times$10 chain each prompt is sent ten times, so the 800-row slice holds 81 distinct prompts. The replayed and fresh g0 rows are separate runs under the same seed, and they differ slightly because vLLM reruns are not bit-identical (Appendix~\ref{app:lottery}).}
\label{tab:Gchains}
\footnotesize
\setlength{\tabcolsep}{3pt}
\begin{adjustbox}{max width=\linewidth}
\begin{tabular}{llcccccccc}
\toprule
chain & gen & u4 & self-BLEU & PPL & words & first-3 & top df & coupled u4 & uncoupled u4 \\
\midrule
StableLM-2-1.6B, $f=0$ (independent) & g0 & 0.999 & 0.000 & 93 & 90.6 & 0.000 & 0.006 & -- & 0.999 \\
 & g1 & 0.999 & 0.000 & 120 & 96.6 & 0.003 & 0.005 & -- & 0.999 \\
 & g2 & 0.999 & 0.000 & 128 & 98.3 & 0.000 & 0.010 & -- & 0.999 \\
 & g3 & 0.999 & 0.001 & 132 & 100.1 & 0.000 & 0.010 & -- & 0.999 \\
\addlinespace
StableLM-2-1.6B, $f=0.1$ & g0 & 0.999 & 0.001 & 91 & 90.6 & 0.006 & 0.005 & 0.997 & 0.999 \\
 & g1 & 0.999 & 0.000 & 118 & 97.2 & 0.003 & 0.006 & 0.997 & 0.999 \\
 & g2 & 0.998 & 0.000 & 123 & 100.4 & 0.003 & 0.008 & 0.993 & 0.999 \\
 & g3 & 0.998 & 0.001 & 132 & 100.6 & 0.000 & 0.006 & 0.991 & 0.999 \\
\addlinespace
StableLM-2-1.6B, $f=0.3$ & g0 & 0.996 & 0.001 & 91 & 91.3 & 0.019 & 0.009 & 0.992 & 0.999 \\
 & g1 & 0.994 & 0.001 & 137 & 96.9 & 0.011 & 0.015 & 0.982 & 0.999 \\
 & g2 & 0.988 & 0.005 & 174 & 97.9 & 0.008 & 0.054 & 0.963 & 0.999 \\
 & g3 & 0.977 & 0.006 & 193 & 99.5 & 0.008 & 0.139 & 0.930 & 1.000 \\
\addlinespace
StableLM-2-1.6B, $f=1$ (replayed) & g0 & 0.981 & 0.009 & 86 & 88.6 & 0.090 & 0.029 & 0.981 & -- \\
 & g1 & 0.904 & 0.045 & 231 & 95.5 & 0.069 & 0.186 & 0.904 & -- \\
 & g2 & 0.437 & 0.455 & 1{,}403 & 95.4 & 0.074 & 0.721 & 0.437 & -- \\
 & g3 & 0.152 & 0.822 & 3{,}179 & 95.5 & 0.118 & 1.000 & 0.152 & -- \\
\addlinespace
StableLM-2-1.6B, fresh seed per generation & g0 & 0.981 & 0.009 & 85 & 89.3 & 0.089 & 0.026 & 0.981 & -- \\
 & g1 & 0.970 & 0.009 & 111 & 100.5 & 0.103 & 0.034 & 0.970 & -- \\
 & g2 & 0.961 & 0.013 & 128 & 101.3 & 0.028 & 0.079 & 0.961 & -- \\
 & g3 & 0.955 & 0.016 & 128 & 102.7 & 0.159 & 0.053 & 0.955 & -- \\
 & g4 & 0.944 & 0.027 & 130 & 103.7 & 0.046 & 0.059 & 0.944 & -- \\
 & g5 & 0.950 & 0.019 & 131 & 105.7 & 0.088 & 0.066 & 0.950 & -- \\
 & g6 & 0.937 & 0.030 & 139 & 107.5 & 0.073 & 0.068 & 0.937 & -- \\
 & g7 & 0.949 & 0.015 & 129 & 109.6 & 0.041 & 0.025 & 0.949 & -- \\
 & g8 & 0.946 & 0.020 & 136 & 111.3 & 0.026 & 0.055 & 0.946 & -- \\
\addlinespace
StableLM-2-1.6B, per-request seed fixed across generations & g0 & 0.999 & 0.000 & 93 & 90.6 & 0.000 & 0.006 & -- & 0.999 \\
 & g1 & 0.999 & 0.000 & 121 & 97.2 & 0.010 & 0.006 & -- & 0.999 \\
 & g2 & 0.997 & 0.000 & 137 & 98.9 & 0.000 & 0.009 & -- & 0.997 \\
 & g3 & 0.998 & 0.000 & 135 & 100.4 & 0.003 & 0.008 & -- & 0.998 \\
\addlinespace
StableLM-2-1.6B, independent, weights chained across generations & g0 & 0.999 & 0.000 & 93 & 90.6 & 0.000 & 0.006 & -- & 0.999 \\
 & g1 & 0.999 & 0.000 & 120 & 96.8 & 0.003 & 0.005 & -- & 0.999 \\
 & g2 & 0.999 & 0.000 & 131 & 100.4 & 0.000 & 0.010 & -- & 0.999 \\
 & g3 & 0.998 & 0.000 & 132 & 101.5 & 0.000 & 0.010 & -- & 0.998 \\
 & g4 & 0.998 & 0.000 & 138 & 101.5 & 0.003 & 0.015 & -- & 0.998 \\
 & g5 & 0.998 & 0.000 & 133 & 103.7 & 0.000 & 0.015 & -- & 0.998 \\
\addlinespace
StableLM-2-1.6B, prompts $\times$10, independent & g0 & 0.998 & 0.000 & 96 & 90.3 & 0.063 & 0.006 & -- & 0.998 \\
 & g1 & 0.998 & 0.001 & 119 & 97.3 & 0.049 & 0.008 & -- & 0.998 \\
 & g2 & 0.999 & 0.000 & 134 & 98.8 & 0.084 & 0.006 & -- & 0.999 \\
 & g3 & 0.999 & 0.000 & 128 & 100.7 & 0.049 & 0.005 & -- & 0.999 \\
\addlinespace
StableLM-2-1.6B, temperature 0.7, independent & g0 & 0.991 & 0.001 & 22 & 92.5 & 0.009 & 0.016 & -- & 0.991 \\
 & g1 & 0.986 & 0.005 & 22 & 99.6 & 0.013 & 0.020 & -- & 0.986 \\
 & g2 & 0.983 & 0.004 & 27 & 104.7 & 0.010 & 0.034 & -- & 0.983 \\
 & g3 & 0.973 & 0.007 & 30 & 106.6 & 0.008 & 0.063 & -- & 0.973 \\
 & g4 & 0.959 & 0.016 & 36 & 108.4 & 0.019 & 0.109 & -- & 0.959 \\
 & g5 & 0.945 & 0.026 & 42 & 110.1 & 0.014 & 0.163 & -- & 0.945 \\
 & g6 & 0.916 & 0.043 & 48 & 111.4 & 0.008 & 0.203 & -- & 0.916 \\
\addlinespace
StableLM-2-1.6B, temperature 0.7, $f=1$ (replayed) & g0 & 0.989 & 0.002 & 22 & 91.7 & 0.068 & 0.019 & 0.989 & -- \\
 & g1 & 0.984 & 0.004 & 22 & 99.3 & 0.035 & 0.016 & 0.984 & -- \\
 & g2 & 0.979 & 0.010 & 26 & 104.0 & 0.028 & 0.025 & 0.979 & -- \\
 & g3 & 0.969 & 0.016 & 30 & 106.4 & 0.014 & 0.045 & 0.969 & -- \\
 & g4 & 0.950 & 0.020 & 34 & 107.4 & 0.008 & 0.059 & 0.950 & -- \\
 & g5 & 0.927 & 0.037 & 40 & 109.1 & 0.023 & 0.101 & 0.927 & -- \\
 & g6 & 0.897 & 0.056 & 45 & 109.8 & 0.020 & 0.178 & 0.897 & -- \\
\addlinespace
StableLM-2-1.6B, $f=1$ (replayed), chain seed 43 & g0 & 0.981 & 0.006 & 94 & 90.7 & 0.078 & 0.018 & 0.981 & -- \\
 & g1 & 0.915 & 0.039 & 216 & 94.3 & 0.134 & 0.105 & 0.915 & -- \\
 & g2 & 0.399 & 0.515 & 1{,}385 & 91.8 & 0.175 & 0.716 & 0.399 & -- \\
 & g3 & 0.070 & 0.908 & 4{,}824 & 89.7 & 0.396 & 1.000 & 0.070 & -- \\
\addlinespace
StableLM-2-1.6B, fresh seed per generation, chain seed 43 & g0 & 0.981 & 0.010 & 94 & 91.3 & 0.079 & 0.018 & 0.981 & -- \\
 & g1 & 0.968 & 0.010 & 114 & 100.4 & 0.095 & 0.028 & 0.968 & -- \\
 & g2 & 0.967 & 0.006 & 143 & 101.0 & 0.111 & 0.026 & 0.967 & -- \\
 & g3 & 0.971 & 0.011 & 136 & 97.5 & 0.033 & 0.030 & 0.971 & -- \\
\bottomrule
\end{tabular}
\end{adjustbox}
\end{table}

\begin{table}[p]
\centering
\caption{The cluster-B chains of the other checkpoints, with the columns of Table~\ref{tab:Gchains} (chain seed 42 unless marked).}
\label{tab:Gothers}
\footnotesize
\setlength{\tabcolsep}{3pt}
\begin{adjustbox}{max totalsize={\linewidth}{0.93\textheight}}
\begin{tabular}{llcccccccc}
\toprule
chain & gen & u4 & self-BLEU & PPL & words & first-3 & top df & coupled u4 & uncoupled u4 \\
\midrule
StableLM-3B, $f=0.3$ & g0 & 0.997 & 0.000 & 119 & 90.1 & 0.014 & 0.010 & 0.992 & 0.999 \\
 & g1 & 0.980 & 0.004 & 179 & 98.5 & 0.014 & 0.044 & 0.940 & 0.999 \\
 & g2 & 0.891 & 0.087 & 275 & 101.3 & 0.011 & 0.159 & 0.655 & 0.999 \\
 & g3 & 0.783 & 0.144 & 464 & 101.8 & 0.020 & 0.359 & 0.373 & 0.976 \\
 & g4 & 0.697 & 0.253 & 576 & 102.6 & 0.035 & 0.453 & 0.261 & 0.905 \\
 & g5 & 0.625 & 0.311 & 695 & 102.4 & 0.061 & 0.504 & 0.200 & 0.826 \\
\addlinespace[2pt]
StableLM-3B, $f=1$ (replayed) & g0 & 0.983 & 0.005 & 121 & 89.2 & 0.089 & 0.019 & 0.983 & -- \\
 & g1 & 0.703 & 0.230 & 863 & 94.8 & 0.128 & 0.251 & 0.703 & -- \\
 & g2 & 0.145 & 0.829 & 5{,}069 & 95.1 & 0.278 & 0.931 & 0.145 & -- \\
 & g3 & 0.051 & 0.939 & 6{,}422 & 94.3 & 0.490 & 0.989 & 0.051 & -- \\
\addlinespace[2pt]
StableLM-3B, fresh seed per generation & g0 & 0.983 & 0.003 & 131 & 89.9 & 0.086 & 0.030 & 0.983 & -- \\
 & g1 & 0.978 & 0.007 & 148 & 102.1 & 0.058 & 0.079 & 0.978 & -- \\
 & g2 & 0.978 & 0.008 & 158 & 103.0 & 0.099 & 0.034 & 0.978 & -- \\
 & g3 & 0.981 & 0.004 & 184 & 104.5 & 0.046 & 0.018 & 0.981 & -- \\
 & g4 & 0.977 & 0.010 & 173 & 105.6 & 0.073 & 0.041 & 0.977 & -- \\
 & g5 & 0.978 & 0.008 & 179 & 106.9 & 0.056 & 0.053 & 0.978 & -- \\
 & g6 & 0.977 & 0.011 & 186 & 108.0 & 0.041 & 0.063 & 0.977 & -- \\
 & g7 & 0.974 & 0.006 & 187 & 109.0 & 0.073 & 0.088 & 0.974 & -- \\
 & g8 & 0.967 & 0.015 & 194 & 109.6 & 0.056 & 0.145 & 0.967 & -- \\
\addlinespace[2pt]
StableLM-3B, per-request seed fixed across generations & g0 & 0.999 & 0.000 & 117 & 90.7 & 0.000 & 0.005 & -- & 0.999 \\
 & g1 & 0.999 & 0.000 & 159 & 98.2 & 0.000 & 0.004 & -- & 0.999 \\
 & g2 & 0.999 & 0.000 & 164 & 100.9 & 0.000 & 0.005 & -- & 0.999 \\
 & g3 & 0.999 & 0.000 & 172 & 103.0 & 0.000 & 0.008 & -- & 0.999 \\
\addlinespace[2pt]
StableLM-3B, temperature 0.7, independent & g0 & 0.995 & 0.001 & 26 & 89.5 & 0.000 & 0.009 & -- & 0.995 \\
 & g1 & 0.990 & 0.001 & 31 & 103.4 & 0.010 & 0.016 & -- & 0.990 \\
 & g2 & 0.982 & 0.005 & 37 & 106.8 & 0.019 & 0.060 & -- & 0.982 \\
 & g3 & 0.975 & 0.008 & 44 & 107.0 & 0.015 & 0.109 & -- & 0.975 \\
 & g4 & 0.963 & 0.019 & 46 & 108.2 & 0.024 & 0.119 & -- & 0.963 \\
 & g5 & 0.952 & 0.016 & 57 & 109.0 & 0.035 & 0.168 & -- & 0.952 \\
 & g6 & 0.938 & 0.024 & 61 & 110.1 & 0.034 & 0.213 & -- & 0.938 \\
\addlinespace[2pt]
StableLM-3B, temperature 0.7, $f=1$ (replayed) & g0 & 0.992 & 0.002 & 25 & 88.1 & 0.035 & 0.013 & 0.992 & -- \\
 & g1 & 0.988 & 0.003 & 32 & 102.1 & 0.023 & 0.026 & 0.988 & -- \\
 & g2 & 0.982 & 0.009 & 37 & 107.0 & 0.039 & 0.075 & 0.982 & -- \\
 & g3 & 0.974 & 0.012 & 45 & 107.9 & 0.049 & 0.105 & 0.974 & -- \\
 & g4 & 0.965 & 0.010 & 52 & 108.6 & 0.058 & 0.136 & 0.965 & -- \\
 & g5 & 0.952 & 0.019 & 59 & 109.3 & 0.064 & 0.170 & 0.952 & -- \\
 & g6 & 0.936 & 0.032 & 64 & 108.9 & 0.070 & 0.151 & 0.936 & -- \\
\addlinespace[2pt]
Minitron-4B, $f=1$ (replayed) & g0 & 0.980 & 0.007 & 116 & 94.1 & 0.134 & 0.015 & 0.980 & -- \\
 & g1 & 0.870 & 0.069 & 361 & 107.1 & 0.103 & 0.140 & 0.870 & -- \\
 & g2 & 0.342 & 0.605 & 1{,}924 & 117.7 & 0.141 & 0.853 & 0.342 & -- \\
 & g3 & 0.117 & 0.860 & 4{,}533 & 120.0 & 0.240 & 1.000 & 0.117 & -- \\
\addlinespace[2pt]
Minitron-4B, fresh seed per generation & g0 & 0.979 & 0.005 & 118 & 93.6 & 0.121 & 0.025 & 0.979 & -- \\
 & g1 & 0.968 & 0.010 & 186 & 102.6 & 0.038 & 0.034 & 0.968 & -- \\
 & g2 & 0.963 & 0.011 & 204 & 111.2 & 0.030 & 0.098 & 0.963 & -- \\
 & g3 & 0.977 & 0.005 & 202 & 113.9 & 0.081 & 0.028 & 0.977 & -- \\
 & g4 & 0.979 & 0.009 & 213 & 114.2 & 0.054 & 0.023 & 0.979 & -- \\
 & g5 & 0.970 & 0.010 & 213 & 113.2 & 0.054 & 0.060 & 0.970 & -- \\
 & g6 & 0.972 & 0.013 & 212 & 115.5 & 0.048 & 0.040 & 0.972 & -- \\
 & g7 & 0.964 & 0.016 & 226 & 114.6 & 0.075 & 0.071 & 0.964 & -- \\
 & g8 & 0.922 & 0.050 & 239 & 117.1 & 0.578 & 0.303 & 0.922 & -- \\
\addlinespace[2pt]
Minitron-4B, $f=0$ (independent), chain seed 43 & g0 & 0.997 & 0.001 & 121 & 93.8 & 0.003 & 0.008 & -- & 0.997 \\
 & g1 & 0.998 & 0.001 & 211 & 98.9 & 0.000 & 0.006 & -- & 0.998 \\
 & g2 & 0.999 & 0.001 & 229 & 103.0 & 0.000 & 0.010 & -- & 0.999 \\
 & g3 & 0.999 & 0.000 & 250 & 108.8 & 0.003 & 0.011 & -- & 0.999 \\
\addlinespace[2pt]
Minitron-4B, $f=1$ (replayed), chain seed 43 & g0 & 0.973 & 0.007 & 117 & 98.1 & 0.043 & 0.028 & 0.973 & -- \\
 & g1 & 0.796 & 0.123 & 558 & 104.8 & 0.090 & 0.471 & 0.796 & -- \\
 & g2 & 0.308 & 0.624 & 2{,}428 & 111.2 & 0.374 & 0.884 & 0.308 & -- \\
 & g3 & 0.064 & 0.908 & 5{,}080 & 110.2 & 0.850 & 0.999 & 0.064 & -- \\
\addlinespace[2pt]
Granite-3.3-2B, $f=0$ (independent), chain seed 43 & g0 & 0.996 & 0.000 & 96 & 79.7 & 0.003 & 0.010 & -- & 0.996 \\
 & g1 & 0.998 & 0.000 & 348 & 78.8 & 0.008 & 0.010 & -- & 0.998 \\
 & g2 & 0.999 & 0.000 & 884 & 74.1 & 0.000 & 0.005 & -- & 0.999 \\
 & g3 & 0.997 & 0.002 & 539 & 65.8 & 0.000 & 0.009 & -- & 0.997 \\
\addlinespace[2pt]
Granite-3.3-2B, $f=1$ (replayed), chain seed 43 & g0 & 0.989 & 0.002 & 95 & 78.5 & 0.075 & 0.018 & 0.989 & -- \\
 & g1 & 0.913 & 0.040 & 540 & 75.7 & 0.123 & 0.078 & 0.913 & -- \\
 & g2 & 0.387 & 0.501 & 1{,}974 & 76.1 & 0.254 & 0.544 & 0.387 & -- \\
 & g3 & 0.068 & 0.898 & 3{,}976 & 81.8 & 0.655 & 0.981 & 0.068 & -- \\
\addlinespace[2pt]
Qwen3-1.7B, $f=1$ (replayed) & g0 & 0.947 & 0.023 & 120 & 88.8 & 0.088 & 0.048 & 0.947 & -- \\
 & g1 & 0.909 & 0.049 & 278 & 96.3 & 0.089 & 0.101 & 0.909 & -- \\
 & g2 & 0.656 & 0.244 & 649 & 99.2 & 0.150 & 0.599 & 0.656 & -- \\
 & g3 & 0.285 & 0.611 & 1{,}389 & 89.7 & 0.314 & 0.928 & 0.285 & -- \\
\addlinespace[2pt]
Qwen3-1.7B, $f=0$ (independent) & g0 & 0.961 & 0.034 & 125 & 86.0 & 0.003 & 0.071 & -- & 0.961 \\
 & g1 & 0.993 & 0.005 & 211 & 93.3 & 0.000 & 0.019 & -- & 0.993 \\
 & g2 & 0.996 & 0.000 & 305 & 91.2 & 0.004 & 0.009 & -- & 0.996 \\
 & g3 & 0.981 & 0.005 & 325 & 82.2 & 0.000 & 0.034 & -- & 0.981 \\
\addlinespace[2pt]
Qwen2.5-1.5B, $f=1$ (replayed) & g0 & 0.930 & 0.049 & 91 & 85.5 & 0.059 & 0.123 & 0.930 & -- \\
 & g1 & 0.928 & 0.068 & 168 & 89.7 & 0.084 & 0.074 & 0.928 & -- \\
 & g2 & 0.770 & 0.152 & 551 & 92.1 & 0.165 & 0.225 & 0.770 & -- \\
 & g3 & 0.280 & 0.622 & 1{,}837 & 78.3 & 0.365 & 0.934 & 0.280 & -- \\
\addlinespace[2pt]
Qwen2.5-1.5B, $f=0$ (independent) & g0 & 0.933 & 0.055 & 85 & 81.2 & 0.003 & 0.139 & -- & 0.933 \\
 & g1 & 0.958 & 0.032 & 138 & 89.8 & 0.000 & 0.079 & -- & 0.958 \\
 & g2 & 0.986 & 0.009 & 236 & 93.7 & 0.003 & 0.034 & -- & 0.986 \\
 & g3 & 0.992 & 0.007 & 300 & 89.4 & 0.005 & 0.029 & -- & 0.992 \\
\addlinespace[2pt]
SmolLM3-3B, $f=1$ (replayed) & g0 & 0.986 & 0.003 & 79 & 94.1 & 0.079 & 0.015 & 0.986 & -- \\
 & g1 & 0.965 & 0.008 & 119 & 99.0 & 0.055 & 0.050 & 0.965 & -- \\
 & g2 & 0.802 & 0.117 & 290 & 100.7 & 0.051 & 0.395 & 0.802 & -- \\
 & g3 & 0.377 & 0.589 & 940 & 100.1 & 0.059 & 0.791 & 0.377 & -- \\
\addlinespace[2pt]
SmolLM3-3B, $f=0$ (independent) & g0 & 0.999 & 0.000 & 79 & 95.8 & 0.000 & 0.005 & -- & 0.999 \\
 & g1 & 0.999 & 0.000 & 103 & 100.2 & 0.005 & 0.008 & -- & 0.999 \\
 & g2 & 0.999 & 0.000 & 112 & 104.4 & 0.003 & 0.006 & -- & 0.999 \\
 & g3 & 0.999 & 0.000 & 117 & 106.4 & 0.000 & 0.011 & -- & 0.999 \\
\bottomrule
\end{tabular}
\end{adjustbox}
\end{table}

\clearpage
\section{The lottery}
\label{app:lottery}

\paragraph{Code path.}
In vLLM 0.10.2, \texttt{v1/worker/gpu\_model\_runner.py} (line 541) gives every seeded request a \texttt{torch.Generator(device="cuda")} seeded with the request's seed. The sampler casts logits to float32, applies the repetition and frequency penalties and the temperature, masks ids outside the top-p nucleus, and \texttt{random\_sample} (\texttt{v1/sample/ops/topk\_topp\_sampler.py}, lines 194 to 215) fills row $i$ with \texttt{q[i].exponential\_(generator=generator\_i)} and returns \texttt{argmax(probs / q)}. The source gives the reason for the race: it avoids the CPU-GPU synchronisation of a multinomial call. Each call advances the generator by a Philox offset of 4. The V0 sampler builds \texttt{torch.Generator(device).manual\_seed(seed)} per sequence group (\texttt{model\_executor/sampling\_metadata.py}) and draws the exponential noise from it in \texttt{\_multinomial}. Reconstructing $q(t)$ for $t=0$ to 127 at the ten vocabulary widths of the panel (32{,}000 to 256{,}000) gives bit-identical values on every id below 32{,}000 at every step, and $q(0)$ differs from $q(1)$.

\paragraph{Replay.}
Rows 0 to 23 of the original g0 streams of StableLM-2-1.6B and SmolLM2-1.7B were replayed with the Hugging Face base models (bf16, SDPA) through the same penalty, temperature and top-p code, with $q(t)$ from \texttt{torch.Generator("cuda").manual\_seed(42)}, resynchronising to the original text after each mismatch. When the replay's top candidate was a strict prefix of the original token, the resynchronisation accepted it and the row left the original path (5 rows of StableLM-2-1.6B and 4 of SmolLM2-1.7B). The right-hand columns of Table~\ref{tab:Freplay} add one backtracking step whenever the best consistent token lies more than 3 nats of $\log p/q$ below the argmax. It needed one backtrack in each derailed row and left the other rows unchanged. Byte-exact replay of a whole row succeeds for 3 of 24 rows (12.5\%), close to vLLM's own rate when it reruns the same prompts with the same seed (13\% for StableLM-2-1.6B, 7\% for SmolLM2-1.7B), because bf16 kernels differ and the two leading candidates of a step are sometimes within a few hundredths of a nat.

\begin{table}[h]
\centering
\caption{Replaying the original generation-0 continuations of StableLM-2-1.6B and SmolLM2-1.7B (24 rows each) with the lottery reconstructed from seed 42. ``Lottery steps'' are steps where the original token is not the most probable after penalties and top-p. The right pair of columns adds one backtracking step to the resynchronisation of the replay tool (Appendix~\ref{app:lottery}).}
\label{tab:Freplay}
\small
\begin{adjustbox}{max width=\linewidth}
\begin{tabular}{lcccc}
\toprule
 & \multicolumn{2}{c}{replay as first run} & \multicolumn{2}{c}{with prefix repair} \\
\cmidrule(lr){2-3}\cmidrule(lr){4-5}
 & StableLM-2-1.6B & SmolLM2-1.7B & StableLM-2-1.6B & SmolLM2-1.7B \\
\midrule
steps where the original token is not the most probable & 0.637 & 0.684 & 0.634 & 0.681 \\
agreement with the original, all steps & 0.903 & 0.872 & 0.982 & 0.977 \\
agreement with the original, lottery steps & 0.890 & 0.858 & 0.979 & 0.971 \\
control: noise of seed 43, all / lottery steps & 0.235 / 0.044 & 0.199 / 0.041 & 0.236 / 0.044 & 0.201 / 0.042 \\
control: noise of step $t{+}1$, all / lottery steps & 0.227 / 0.042 & 0.207 / 0.042 & 0.228 / 0.043 & 0.203 / 0.040 \\
rows reproduced byte for byte & 3/24 & 3/24 & 3/24 & 3/24 \\
rows that left the original path & 5 & 4 & 0 & 0 \\
median $q$-quantile of the original token, lottery steps & 0.0123 & 0.0101 & 0.0076 & 0.0070 \\
\bottomrule
\end{tabular}
\end{adjustbox}
\end{table}

\paragraph{Step-locking.}
An (id, step) pair is step-locked when the id is written at that step in at least 8 of 800 continuations and at least four times as often as at the other steps. The thirteen g0 streams hold 874 to 1{,}677 such pairs each, 16{,}151 in total and spread over all 127 steps. The same rule finds 3 to 23 pairs in the independent-seed regenerations of Qwen3-1.7B, Qwen2.5-1.5B, SmolLM2-1.7B, SmolLM3-3B and StableLM-2-1.6B. At step 14, the id for `` in'' is written into 262 of StableLM-2-1.6B's 800 continuations against about 5 expected from its rate at other steps, and its draw lies at the 0.13\% quantile of that step. The same id at the same step is written into 281 of SmolLM3-3B's and 268 of OLMo-2-1B's continuations, which share StableLM-2-1.6B's id map, and into 1 of StableLM-3B's and 2 of SmolLM2-1.7B's, whose maps spell it differently. Table~\ref{tab:Fdose} bins frequent ids by the quantile of their draw, and Table~\ref{tab:Fcolock} counts co-locks between pairs of checkpoints. In the race, an id with probability 0.05 in many contexts is written nearly eight times as often as its base rate when its draw sits at the 5\% quantile, so locking reaches well past the lowest 1\% of draws.

\begin{table}[h]
\centering
\caption{Which ids lock. Frequent ids (100 or more occurrences) at steps 1 to 127 of the thirteen generation-0 streams, binned by the quantile of the exponential draw $q$ they receive at that step, and the share of these id-steps that are step-locked (the id is written at that step in at least 8 of 800 continuations and at four times its rate at other steps). The table covers the 7{,}122 step-locked pairs of frequent ids among the 16{,}151 that the thirteen streams hold under the shared seed (Appendix~\ref{app:lottery}).}
\label{tab:Fdose}
\small
\begin{adjustbox}{max width=\linewidth}
\begin{tabular}{lccccc}
\toprule
q-quantile at step $t$ & lowest 0.1\% & 0.1\% to 1\% & 1\% to 3\% & 3\% to 10\% & above 10\% \\
\midrule
id-steps & 429 & 2{,}153 & 4{,}346 & 14{,}908 & 191{,}016 \\
share step-locked & 0.953 & 0.778 & 0.436 & 0.123 & 0.007 \\
\bottomrule
\end{tabular}
\end{adjustbox}
\end{table}

\begin{table}[h]
\centering
\caption{Step-locked (id, step) pairs shared by two checkpoints at generation 0 under the same seed. Off-step: mean number of co-locks when one stream is shifted by 1 to 3 steps.}
\label{tab:Fcolock}
\small
\begin{adjustbox}{max width=\linewidth}
\begin{tabular}{lcccc}
\toprule
pair & locked pairs & same-step co-locks & off-step & Jaccard \\
\midrule
Qwen3-1.7B, Qwen2.5-1.5B (Qwen map) & 1{,}196 / 998 & 743 & 26.2 & 0.512 \\
SmolLM3-3B, OLMo-2-1B (cl100k core) & 1{,}100 / 874 & 631 & 18.3 & 0.470 \\
SmolLM3-3B, StableLM-2-1.6B (cl100k core) & 1{,}100 / 1{,}143 & 770 & 20.8 & 0.523 \\
OLMo-2-1B, StableLM-2-1.6B (cl100k core) & 874 / 1{,}143 & 628 & 15.7 & 0.452 \\
Qwen2.5-1.5B, StableLM-2-1.6B (largest other pair) & 998 / 1{,}143 & 194 & 6.3 & 0.100 \\
StableLM-2-1.6B, StableLM-3B (same family, different map) & 1{,}143 / 1{,}238 & 151 & 3.2 & 0.068 \\
mean of the 74 pairs without a shared map & & & & 0.046 \\
\bottomrule
\end{tabular}
\end{adjustbox}
\end{table}

\paragraph{Birth of dominant phrases.}
For each seed-42 self-loop under the replayed rule, the ten most frequent 4-grams of the g3 slice with document frequency of at least 5\% were traced back to the first generation in which at least five samples carry them. At that first appearance, 41\% to 97\% of their occurrences lie within three words of the phrase's median position across the twelve chains that have such phrases (SmolLM2-1.7B has none), against about 7\% for random placement. Dominant phrases are born step-locked in fast- and slow-collapsing chains alike.

\section{Dominant phrases}
\label{app:phrases}

Table~\ref{tab:dominant} lists the ten most frequent word 4-grams of every generation in three chains, and Table~\ref{tab:dominanttop1} gives the most frequent one in further chains of Tables~\ref{tab:Gchains} and~\ref{tab:Gothers}. Table~\ref{tab:trace} gives the values plotted in Figure~\ref{fig:resonance}c,d, and Table~\ref{tab:freshlong} the two most frequent 4-grams at g4 to g8 of the three fresh chains followed past g3.

Checkpoints that share an id map grow the same phrases under the replayed seed. The 3-gram ``La County Public'' is in 35\% of StableLM-2-1.6B's samples at g2 and in 71\% of SmolLM3-3B's at g3, and the 2-gram ``car wash'' in 67\% of Qwen3-1.7B's at g2 and in 93\% of Qwen2.5-1.5B's at g3. Neither phrase occurs in the independent-seed chains of these four checkpoints.

Over the fresh chains followed to g8, the most frequent phrase of StableLM-2-1.6B changes from one generation to the next and stays under 8\% of samples, and that of Minitron-4B changes through g5 and holds 2\% to 10\% of samples through g7 (Tables~\ref{tab:dominanttop1}, \ref{tab:freshlong} and~\ref{tab:Gothers}). In StableLM-3B the phrase ``British Army Chief General'' is absent from g0 to g3, enters at g4 in second place with 3.4\% of samples, and leads from g5 on, its share growing in every generation to 14.5\% at g8. In Minitron-4B the most frequent 4-gram jumps from 7.1\% of samples at g7 to 30.3\% at g8, and the most common three-word opening from at most 1.4\% in any earlier generation to 30.1\%, which fits an opening written at the first steps, where every request meets the same draws. At g8 the ten most frequent 4-grams of this chain sit within three words of their median position in 95\% of their first occurrences, against 27\% for the five of StableLM-3B and about 7\% for random placement (the index of Appendix~\ref{app:lottery}).

Under partial coupling a phrase fills the coupled rows before it reaches the others (Table~\ref{tab:split}). The uncoupled rows of a generation carry a phrase at a rate that rises with its share of the pool that trained their model, far below that share while the phrase is rare in the pool and at half to three quarters of it once the phrase holds 15\% of the pool or more. The coupled rows carry a phrase at many times its share from the generation after it first appears, which is how a phrase becomes common in the pool before the uncoupled rows take it up.

\IfFileExists{tables/qualitative_trace.tex}{\begin{table*}[t]
\centering\scriptsize
\setlength{\tabcolsep}{4pt}
\caption{Verbatim excerpts from the StableLM-2-1.6B chain (seed 42), one per generation, each a window of about 14 words. Left, replayed seed: the marked phrase is the most frequent word 4-gram at g3, with the number of that generation's 800 measured samples that contain it. Right, fresh seed per generation: the marked phrase is that generation's own most frequent 4-gram, with its count.}
\label{tab:qualitative}
\begin{tabular}{c p{0.42\linewidth} p{0.42\linewidth}}
\toprule
gen & replayed seed & fresh seed per generation \\
\midrule
g0 & ...~Zoe College Poland - Requiem \texttt{Directory Wedding Simple Instructions} Matthew Hayden With Dee Full~... \newline\textit{in 2 of 800 samples} & ...~interesting response, maybe because it's \texttt{originally intended for only} French readers is this: For~... \newline\textit{in 21 of 800 samples, in no other generation} \\
\addlinespace[3pt]
g1 & ...~College Poland Twitter Reagents Writing \texttt{Directory Wedding Simple Instructions} Matthew Hayden Royal Geographical Society~... \newline\textit{in 16 of 800 samples} & ...~authentic version designed so and \texttt{as far I know} today only Park Güell is~... \newline\textit{in 27 of 800 samples, in no other generation} \\
\addlinespace[3pt]
g2 & ...~College Poland Twitter Reagents give \texttt{Directory Wedding Simple Instructions} business warehouse Royal Geographical Society~... \newline\textit{in 577 of 800 samples} & ...~training time raised my opinion \texttt{. Who knows? They} say if your future mother-in-law~... \newline\textit{in 63 of 800 samples, in no other generation} \\
\addlinespace[3pt]
g3 & ...~College Poland Twitter Reagents restoration \texttt{Directory Wedding Simple Instructions} Spitfire Royal Gunnery Brothers Thurston~... \newline\textit{in 800 of 800 samples} & ...~Black Widow @ Ophidian Metallophilia. \texttt{However I wasn't aware} that it existed until just~... \newline\textit{in 42 of 800 samples, in no other generation} \\
\addlinespace[3pt]
\bottomrule
\end{tabular}
\end{table*}
}{\begin{table}[t]\centering\fbox{\parbox[c][1.5in][c]{0.9\linewidth}{\centering draft: \texttt{tables/qualitative\_trace.tex}}}\caption{Qualitative trace.}\label{tab:qualitative}\end{table}}
\begin{table}[h]
\centering
\caption{The values plotted in Figure~\ref{fig:resonance}c,d: document frequency (\% of the 800-sample slice) at g0 to g3 of the phrase that leads each generation of the StableLM-2-1.6B chain (seed 42), under the replayed and the fresh rule. Lead gen: the first generation in which the phrase is the most frequent word 4-gram.}
\label{tab:trace}
\scriptsize
\begin{adjustbox}{max width=\linewidth}
\begin{tabular}{lllcccc}
\toprule
seed rule & lead gen & phrase & g0 & g1 & g2 & g3 \\
\midrule
replayed & g0 & originally intended for only & 2.9 & 6.1 & 2.9 & 0.0 \\
 & g1 & La County Public Library & 1.3 & 18.6 & 25.9 & 0.8 \\
 & g2 & Directory Wedding Simple Instructions & 0.3 & 2.0 & 72.1 & 100.0 \\
\addlinespace
fresh & g0 & originally intended for only & 2.6 & 0.0 & 0.0 & 0.0 \\
 & g1 & as far I know & 0.0 & 3.4 & 0.0 & 0.0 \\
 & g2 & . Who knows? They & 0.0 & 0.0 & 7.9 & 0.0 \\
 & g3 & However I wasn't aware & 0.0 & 0.0 & 0.0 & 5.3 \\
\bottomrule
\end{tabular}
\end{adjustbox}
\end{table}

\begin{table}[p]
\centering
\caption{Top ten word 4-grams per generation with their document frequency (\% of the 800-sample slice) in three chains. Under the replayed seed the phrases persist and grow into a template. Under a fresh seed per generation each generation's list shares nothing with the next.}
\label{tab:dominant}
\scriptsize
\begin{adjustbox}{max width=\linewidth}
\begin{tabular}{llp{0.74\linewidth}}
\toprule
chain & gen & top ten 4-grams (document frequency, \%) \\
\midrule
StableLM-2-1.6B, replayed seed & g0 & \parbox[t]{0.74\linewidth}{\raggedright originally intended for only (2.9); must be logged in (1.5); You must be logged (1.5); . You must be (1.5); be logged in to (1.4); logged in to post (1.3); La County Public Library (1.3); to post a comment (1.1); in to post a (1.1); I became closer to (1.0)} \\[2pt]
 & g1 & \parbox[t]{0.74\linewidth}{\raggedright La County Public Library (18.6); originally intended for only (6.1); strategy facilitates only through (4.6); postal address for correspondence (4.3); - much changed or (4.3); can't do anything about (4.0); Tariff keys are entitled (4.0); looked at La County (4.0); used La County Public (4.0); ``He can't do anything (3.9)} \\[2pt]
 & g2 & \parbox[t]{0.74\linewidth}{\raggedright Directory Wedding Simple Instructions (72.1); - Zoe College Poland (71.4); Thurston Quentin Diogenes Pack (66.8); Brothers Thurston Quentin Diogenes (61.3); La County Federation AFL (59.0); Wedding Simple Instructions Spitfire (51.4); Simple Instructions Spitfire Royal (51.0); Zoe College Poland Twitter (45.1); restoration Directory Wedding Simple (39.8); College Poland Twitter Reagents (37.8)} \\[2pt]
 & g3 & \parbox[t]{0.74\linewidth}{\raggedright Directory Wedding Simple Instructions (100.0); La County Federation AFL (99.9); Simple Instructions Spitfire Royal (99.9); Wedding Simple Instructions Spitfire (99.9); restoration Directory Wedding Simple (99.9); Mensighanya La County Federation (98.8); Pack Mensighanya La County (98.8); Diogenes Pack Mensighanya La (98.8); Quentin Diogenes Pack Mensighanya (98.8); Thurston Quentin Diogenes Pack (98.6)} \\[2pt]
\midrule
StableLM-2-1.6B, fresh seed per generation & g0 & \parbox[t]{0.74\linewidth}{\raggedright originally intended for only (2.6); La County Public Library (1.6); must be logged in (1.5); You must be logged (1.5); . You must be (1.5); be logged in to (1.4); logged in to post (1.3); to post a comment (1.1); in to post a (1.1); it meant so much (1.0)} \\[2pt]
 & g1 & \parbox[t]{0.74\linewidth}{\raggedright as far I know (3.4); and as far away (3.1); and as far I (2.6); as far away as (2.3); as far away forward (2.0); In addition did this (1.6); . The majority could (1.4); so and as far (1.4); as far back as (1.4); did this. However while (1.4)} \\[2pt]
 & g2 & \parbox[t]{0.74\linewidth}{\raggedright . Who knows? They (7.9); . Who knows? Will (4.6); Who knows? They may (4.3); knows? They say if (3.3); Who knows? They say (3.3); Yours truly would have (2.8); truly would also like (2.3); Yours truly would also (2.3); Who knows? Will there (2.1); truly would have said (2.0)} \\[2pt]
 & g3 & \parbox[t]{0.74\linewidth}{\raggedright However I wasn't aware (5.3); I wasn't aware that (5.0); wasn't aware that it (4.8); when it came up (2.6); As far back as (2.5); when it came time (2.5); aware that it existed (2.3); many will argue that (2.1); that it existed until (2.1); will argue that nothing (2.0)} \\[2pt]
\midrule
StableLM-3B, replayed seed & g0 & \parbox[t]{0.74\linewidth}{\raggedright -- which may be (1.9); however in order for (1.5); which may be maintained (1.4); in order to give (1.4); -- which may not (1.3); Seductive Rooms at El (1.3); order to give these (1.0); may not have been (0.9); all information will be (0.9); pm - 5:50 pm (0.9)} \\[2pt]
 & g1 & \parbox[t]{0.74\linewidth}{\raggedright -- Price Check street (25.1); however in LaCava these (23.5); -- Price guide street (22.8); in LaCava these terms (20.4); LaCava Will -- Price (18.5); ``pants'' without command delay (18.1); in LaCava Will -- (16.6); Will -- Price Check (16.3); without command delay single (14.8); strength ``pants'' without command (14.5)} \\[2pt]
 & g2 & \parbox[t]{0.74\linewidth}{\raggedright \textasciigrave{} half ready pay (93.1); single stepped foremost Captain (91.3); -- continued deliberate Breach (88.8); market maintained No tricks (87.4); spending proper forward temporary (87.1); welcome all Volume III (86.8); Best \textasciigrave{} half ready (85.4); Show Much Committee will (81.5); Committee will -- however (77.1); all Volume III Show (76.8)} \\[2pt]
 & g3 & \parbox[t]{0.74\linewidth}{\raggedright market maintained No tricks (98.9); spending proper forward temporary (98.8); \textasciigrave{} half ready pay (98.5); Committee will -- however (98.5); single stepped foremost Captain (98.3); Best \textasciigrave{} half ready (98.0); stepped foremost Captain Birney (97.9); ``pants'' Palmer Variations single (97.6); strength ``pants'' Palmer Variations (97.6); Show Much Committee will (97.6)} \\[2pt]
\bottomrule
\end{tabular}
\end{adjustbox}
\end{table}

\begin{table}[h]
\centering
\caption{The most frequent word 4-gram per generation (document frequency, \%) in the partial-coupling, trigger, replication and fixed per-request chains, and in the replayed chains of three further checkpoints. ``Replayed'' reuses the chain seed in every generation, ``fresh'' draws a new shared seed per generation, and ``own seed fixed per request'' gives each request its own seed, the same in every generation. SmolLM3-3B shares StableLM-2-1.6B's id map and Qwen3-1.7B shares Qwen2.5-1.5B's.}
\label{tab:dominanttop1}
\scriptsize
\begin{adjustbox}{max width=\linewidth}
\begin{tabular}{lllll}
\toprule
chain & g0 & g1 & g2 & g3 \\
\midrule
StableLM-2-1.6B, $f=0.3$ & \parbox[t]{0.19\linewidth}{\raggedright originally intended for only (0.9)} & \parbox[t]{0.19\linewidth}{\raggedright originally intended for only (1.5)} & \parbox[t]{0.19\linewidth}{\raggedright La County Public Access (5.4)} & \parbox[t]{0.19\linewidth}{\raggedright La County Public Access (13.9)} \\[2pt]
StableLM-2-1.6B, prompts $\times$10 & \parbox[t]{0.19\linewidth}{\raggedright when it comes to (0.6)} & \parbox[t]{0.19\linewidth}{\raggedright during World War II (0.8)} & \parbox[t]{0.19\linewidth}{\raggedright can we conclude that (0.6)} & \parbox[t]{0.19\linewidth}{\raggedright a press conference held (0.5)} \\[2pt]
StableLM-2-1.6B, temperature 0.7, independent & \parbox[t]{0.19\linewidth}{\raggedright one of the most (1.6)} & \parbox[t]{0.19\linewidth}{\raggedright In addition to being (2.0)} & \parbox[t]{0.19\linewidth}{\raggedright In addition to being (3.4)} & \parbox[t]{0.19\linewidth}{\raggedright In addition to being (6.3)} \\[2pt]
StableLM-3B, $f=0.3$ & \parbox[t]{0.19\linewidth}{\raggedright -- which may be (1.0)} & \parbox[t]{0.19\linewidth}{\raggedright -- Price Checkers Later (4.4)} & \parbox[t]{0.19\linewidth}{\raggedright artist Nathan Ranay captured (15.9)} & \parbox[t]{0.19\linewidth}{\raggedright artist Nathan Ranay captured (35.9)} \\[2pt]
StableLM-3B, fresh & \parbox[t]{0.19\linewidth}{\raggedright -- which may be (3.0)} & \parbox[t]{0.19\linewidth}{\raggedright . In order to (7.9)} & \parbox[t]{0.19\linewidth}{\raggedright Since then it has (3.4)} & \parbox[t]{0.19\linewidth}{\raggedright . In order to (1.8)} \\[2pt]
Minitron-4B, replayed & \parbox[t]{0.19\linewidth}{\raggedright Answer: How many days (1.5)} & \parbox[t]{0.19\linewidth}{\raggedright chance ordered those types (14.0)} & \parbox[t]{0.19\linewidth}{\raggedright how busy working singles (85.3)} & \parbox[t]{0.19\linewidth}{\raggedright how busy working singles (100.0)} \\[2pt]
Minitron-4B, fresh & \parbox[t]{0.19\linewidth}{\raggedright Answer: How many days (2.5)} & \parbox[t]{0.19\linewidth}{\raggedright may look like things (3.4)} & \parbox[t]{0.19\linewidth}{\raggedright have provided something different (9.8)} & \parbox[t]{0.19\linewidth}{\raggedright Answer: How many days (2.8)} \\[2pt]
StableLM-2-1.6B, replayed, seed 43 & \parbox[t]{0.19\linewidth}{\raggedright Is Nothing A Longer (1.8)} & \parbox[t]{0.19\linewidth}{\raggedright Despite record support from (10.5)} & \parbox[t]{0.19\linewidth}{\raggedright magnum serviced plus any (71.6)} & \parbox[t]{0.19\linewidth}{\raggedright powerful automotive German military (100.0)} \\[2pt]
StableLM-2-1.6B, fresh, seed 43 & \parbox[t]{0.19\linewidth}{\raggedright Is Nothing A Longer (1.8)} & \parbox[t]{0.19\linewidth}{\raggedright -- but due to (2.8)} & \parbox[t]{0.19\linewidth}{\raggedright For more information call (2.6)} & \parbox[t]{0.19\linewidth}{\raggedright , together with two (3.0)} \\[2pt]
StableLM-2-1.6B, own seed fixed per request & \parbox[t]{0.19\linewidth}{\raggedright in the United States (0.6)} & \parbox[t]{0.19\linewidth}{\raggedright during World War II. (0.6)} & \parbox[t]{0.19\linewidth}{\raggedright It's impossible to say (0.9)} & \parbox[t]{0.19\linewidth}{\raggedright Does it follow that (0.8)} \\[2pt]
StableLM-3B, own seed fixed per request & \parbox[t]{0.19\linewidth}{\raggedright at the end of (0.5)} & \parbox[t]{0.19\linewidth}{\raggedright as part of their (0.4)} & \parbox[t]{0.19\linewidth}{\raggedright during World War II (0.5)} & \parbox[t]{0.19\linewidth}{\raggedright during World War II (0.8)} \\[2pt]
SmolLM3-3B, replayed & \parbox[t]{0.19\linewidth}{\raggedright originally intended for only (1.5)} & \parbox[t]{0.19\linewidth}{\raggedright - much changed with (5.0)} & \parbox[t]{0.19\linewidth}{\raggedright La County Public Access (39.5)} & \parbox[t]{0.19\linewidth}{\raggedright Thurston Quentin performed office (79.1)} \\[2pt]
Qwen3-1.7B, replayed & \parbox[t]{0.19\linewidth}{\raggedright It's impossible to say; (4.8)} & \parbox[t]{0.19\linewidth}{\raggedright It seems like there (10.1)} & \parbox[t]{0.19\linewidth}{\raggedright car wash laundry businesses (59.9)} & \parbox[t]{0.19\linewidth}{\raggedright To save French cuisine (92.8)} \\[2pt]
Qwen2.5-1.5B, replayed & \parbox[t]{0.19\linewidth}{\raggedright paraphrases of each other? (12.3)} & \parbox[t]{0.19\linewidth}{\raggedright paraphrases of each other? (7.4)} & \parbox[t]{0.19\linewidth}{\raggedright ever work car wash (22.5)} & \parbox[t]{0.19\linewidth}{\raggedright keckNow Ronke roll words (93.4)} \\[2pt]
\bottomrule
\end{tabular}
\end{adjustbox}
\end{table}

\begin{table}[h]
\centering
\caption{The two most frequent word 4-grams at g4 to g8 (document frequency, \%) in the three fresh-rule chains followed past g3 (chain seed 42). From g6 on, the Minitron-4B cells give the document frequency of the most frequent 4-gram only. Generations g0 to g3 of these chains are in Tables~\ref{tab:dominant} and~\ref{tab:dominanttop1}.}
\label{tab:freshlong}
\scriptsize
\begin{adjustbox}{max width=\linewidth}
\begin{tabular}{llll}
\toprule
gen & StableLM-2-1.6B & StableLM-3B & Minitron-4B \\
\midrule
g4 & \parbox[t]{0.27\linewidth}{\raggedright . Who knows what (5.9); Only two weeks ago (4.1)} & \parbox[t]{0.27\linewidth}{\raggedright followed shortly thereafter British (4.1); British Army Chief General (3.4)} & \parbox[t]{0.27\linewidth}{\raggedright Answer: How many years (2.3); really matters about this (1.6)} \\[2pt]
g5 & \parbox[t]{0.27\linewidth}{\raggedright Who knows maybe in (6.6); knows maybe in years (3.9)} & \parbox[t]{0.27\linewidth}{\raggedright British Army Chief General (5.3); . In order to (2.8)} & \parbox[t]{0.27\linewidth}{\raggedright More {\fontencoding{T1}\selectfont\guillemotright} Answer: What (6.0); {\fontencoding{T1}\selectfont\guillemotright} Answer: What type (5.3)} \\[2pt]
g6 & \parbox[t]{0.27\linewidth}{\raggedright : while they are (6.8); No matter how close (4.4)} & \parbox[t]{0.27\linewidth}{\raggedright British Army Chief General (6.3); Army Chief General Sir (4.0)} & \parbox[t]{0.27\linewidth}{\raggedright (4.0)} \\[2pt]
g7 & \parbox[t]{0.27\linewidth}{\raggedright we'll probably hear another (2.5); ? Maybe if someone (2.5)} & \parbox[t]{0.27\linewidth}{\raggedright British Army Chief General (8.8); followed shortly thereafter British (6.9)} & \parbox[t]{0.27\linewidth}{\raggedright (7.1)} \\[2pt]
g8 & \parbox[t]{0.27\linewidth}{\raggedright . We will keep (5.5); We will keep them (4.5)} & \parbox[t]{0.27\linewidth}{\raggedright British Army Chief General (14.5); Army Chief General Sir (10.6)} & \parbox[t]{0.27\linewidth}{\raggedright (30.3)} \\[2pt]
\bottomrule
\end{tabular}
\end{adjustbox}
\end{table}

\begin{table}[h]
\centering
\caption{Share (\%) of the coupled slice rows, of the uncoupled slice rows and of the whole 2{,}100-row pool that contain the dominant phrases of the two chains at $f=0.3$, given as coupled / uncoupled / pool. The pool of generation $g$ fine-tunes the model of generation $g{+}1$. The StableLM-3B chain is continued to g5.}
\label{tab:split}
\scriptsize
\begin{adjustbox}{max width=\linewidth}
\begin{tabular}{llcccccc}
\toprule
checkpoint & phrase & g0 & g1 & g2 & g3 & g4 & g5 \\
\midrule
StableLM-2-1.6B & La County Public Access & 0 / 0 / 0 & 3 / 0 / 1 & 17 / 0 / 5 & 43 / 0 / 14 & -- & -- \\
StableLM-3B & artist Nathan Ranay captured & 0 / 0 / 0 & 8 / 0 / 2 & 49 / 0 / 15 & 94 / 8 / 33 & 91 / 23 / 43 & 34 / 31 / 31 \\
StableLM-3B & Committee will interview three & 1 / 0 / 0 & 5 / 0 / 1 & 24 / 0 / 7 & 79 / 2 / 26 & 97 / 16 / 40 & 98 / 28 / 49 \\
\bottomrule
\end{tabular}
\end{adjustbox}
\end{table}

\section{Repeated prompts and temperature 0.7}
\label{sec:nontriggers}

Repeated prompts do not start the loop. Sending each prompt ten times under independent seeds gives StableLM-2-1.6B an opening overlap at g0 of the same order as the shared seed's, yet the chain stays above 0.99 at g3 (Table~\ref{tab:Gchains}). The copies repeat text within a prompt and favour no token at a fixed step across unrelated contexts, the correlation a shared stream supplies.

At temperature 0.7 the replayed seed no longer drives the loop, although the draw stays shared. Its lottery still makes 7\% and 3.5\% of the g0 samples of StableLM-2-1.6B and StableLM-3B share their opening, against 0.9\% and none under independent seeds, yet through six generations each replayed chain tracks the independent-seed chain of the same checkpoint: at g6 the two chains of a checkpoint lie within 0.02 in u4, and the top phrase holds a larger share under independent seeds than under the replayed seed (Tables~\ref{tab:Gchains} and~\ref{tab:Gothers}). Both chains lose lexical diversity slowly, and at about the same rate, so what is lost at this temperature is lost without the shared draw. One reading is that sharpening leaves fewer ids with moderate probability across unrelated contexts, the ids one small draw writes into many samples, so each lottery gives the next fine-tune nothing to amplify. Within six generations the loop is absent at this temperature, not slowed.

\section{The published call pattern}
\label{app:published}

The sampling script of the pipeline, \nolinkurl{src/sample_language_model/sample_language_model.py} in the repository \nolinkurl{RylanSchaeffer/KoyejoLab-Collapse-or-Thrive}, builds \texttt{num\_prompts\_per\_sampling\_call} identical empty prompts and calls \texttt{llm.generate} with one \texttt{SamplingParams(n=num\_samples\_per\_prompt, seed=batch\_generation\_idx, temperature=1.0)}, pinning \texttt{vllm==0.5.4}. The sweep configuration of the released Gemma-2-27B runs, \nolinkurl{sweeps/sample_language_model/accumulate/gemma_2_27b/}, sets both \texttt{num\_prompts\_per\_sampling\_call} and \texttt{num\_samples\_per\_prompt} to 64. We executed the call with 64 prompts and $n=64$, using Qwen2.5-0.5B as a stand-in (bf16, 64 new tokens, top-p 1.0, seed 7). Qwen2.5 has no BOS token, and vLLM 0.5.4 rejects an empty token list, so each empty prompt was replaced by the single token 151643 (\texttt{<|endoftext|>}), one identical special token per request, as Gemma-2's BOS is. Under 0.5.4 all 64 requests return the same 64 sequences in the same child order, byte for byte, and the result is bit-identical on a second GPU. Under 0.10.2 child $k$ of every request receives the same derived seed, 95\% of pairs are byte-identical, and the rest diverge late through batch-dependent numerics. Across calls the seed changes, so each call contributes 64 distinct samples. Table~\ref{tab:H} gives every executed call with the share of steps at which child $k$ of two different requests emits the same token, which is 1.000 for identical prompts under 0.5.4 and 0.024 for distinct prompts under one seed, against 0.006 without a seed.

The released datasets sit under the Hugging Face account \texttt{RylanSchaeffer} with names of the form
\begin{center}\footnotesize\texttt{collapse\_gemma-2-27b\_hs2\_accumulate\_iter\textit{i}\_sftsd\textit{s}\_temp1\_max\_seq\_len512}\end{center}
for iteration $i$ and SFT seed $s$. Iteration 1 with seed 0 and iteration 2 with seeds 0 and 1 were recounted, with 12{,}531 rows each. Duplicate rows make up 49\% of iteration 1 and 93\% and 89\% of iteration 2 with seeds 0 and 1. At iteration 2 with seed 0 the 907 distinct responses fall into 167 responses repeated 54 to 60 times (9{,}555 rows), 60 repeated 27 or 28 times, 222 repeated exactly four times, and 458 singletons. The multiplicities of the first group are consistent with duplication by the call pattern, and those of the other three are not.

\begin{table}[t]
\centering
\caption{The seeding pattern in the pipeline of \citet{kazdan2024collapse}. Rows 1 to 4 execute its call pattern (64 prompts, one \texttt{SamplingParams(n=64, seed=k)} per call, 4{,}096 samples) with Qwen2.5-0.5B as a stand-in model. Shared openings: share of samples whose first three words start another sample. Rows 5 to 7 recount the responses of the pipeline's public Gemma-2-27B datasets. Duplicate rate: 1 minus distinct over total. Repository, file and dataset names are in Appendix~\ref{app:published}.}
\label{tab:wild}
\small
\setlength{\tabcolsep}{4pt}
\begin{tabular}{llrrl}
\toprule
pipeline step & vLLM & distinct / total & shared openings & source \\
\midrule
identical empty prompts, one seed & 0.5.4 & 64 / 4{,}096 & 100\% & executed \\
identical empty prompts, one seed & 0.10.2 & 166 / 4{,}096 & 99\% & executed \\
distinct prompts, one seed & 0.5.4 & 4{,}080 / 4{,}096 & 25\% & executed \\
distinct prompts, no seed & 0.5.4 & 4{,}085 / 4{,}096 & 9\% & executed \\
\addlinespace
released data, iteration 1, SFT seed 0 & 0.5.4 (pinned) & 6{,}422 / 12{,}531 & & recounted \\
released data, iteration 2, SFT seed 0 & 0.5.4 (pinned) & 907 / 12{,}531 & & recounted \\
released data, iteration 2, SFT seed 1 & 0.5.4 (pinned) & 1{,}405 / 12{,}531 & & recounted \\
\bottomrule
\end{tabular}
\end{table}

\begin{table}[h]
\centering
\caption{The published call pattern executed under two vLLM versions (Qwen2.5-0.5B, 64 new tokens, temperature 1). Step-locked agreement: share of positions $t$ at which child $k$ of two different requests emits the same token (baseline about 0.006).}
\label{tab:H}
\small
\setlength{\tabcolsep}{4pt}
\begin{adjustbox}{max width=\linewidth}
\begin{tabular}{llccc}
\toprule
vLLM & call & distinct / total & first-3 shared & step-locked agreement \\
\midrule
0.5.4 & 64 identical one-token prompts, $n{=}64$, seed 7 & 64 / 4{,}096 & 1.000 & 1.000 \\
0.5.4 & 64 $\times$ ``The'', $n{=}64$, seed 7 & 64 / 4{,}096 & 1.000 & 1.000 \\
0.5.4 & 64 distinct prompts, $n{=}64$, seed 7 & 4{,}080 / 4{,}096 & 0.246 & 0.024 \\
0.5.4 & 64 distinct prompts, $n{=}64$, no seed & 4{,}085 / 4{,}096 & 0.090 & 0.006 \\
\addlinespace
0.10.2 & 64 identical one-token prompts, $n{=}64$, seed 7 & 166 / 4{,}096 & 0.994 & 0.973 \\
0.10.2 & 64 $\times$ ``The'', $n{=}64$, seed 7 & 68 / 4{,}096 & 1.000 & 0.999 \\
0.10.2 & 64 distinct prompts, $n{=}64$, seed 7 & 4{,}085 / 4{,}096 & 0.219 & 0.028 \\
0.10.2 & 64 distinct prompts, $n{=}64$, no seed & 4{,}084 / 4{,}096 & 0.091 & 0.006 \\
\bottomrule
\end{tabular}
\end{adjustbox}
\end{table}

\FloatBarrier
\section{Seed code}
\label{app:repro}

The four rules of \S\ref{sec:setup}, for chain seed \texttt{s}, generation \texttt{g} and request index \texttt{i}:

\noindent\begin{minipage}{\linewidth}
\begin{verbatim}
# replayed: one object, the same seed in every generation
out = llm.generate(prompts, SamplingParams(seed=s, **kw))
# fresh: one object, a new seed in every generation
out = llm.generate(prompts, SamplingParams(seed=s + 1000*g, **kw))
# independent: one seed per request
sp = [SamplingParams(seed=s*10**6 + g*10**5 + i, **kw)
      for i in range(len(prompts))]
out = llm.generate(prompts, sp)
# fixed per-request: request i keeps one seed in every generation
sp = [SamplingParams(seed=s*10**6 + i, **kw)
      for i in range(len(prompts))]
out = llm.generate(prompts, sp)
\end{verbatim}
\end{minipage}

For request indices below $10^5$ and generations below 10, the per-request seeds of chain seed $s$ lie in $[s\cdot10^6,(s+1)\cdot10^6)$, so chains with different chain seeds never share a request seed, and the fixed per-request seeds are the independent seeds of g0. Ecosystem members add $10^8$ times their panel number, so their seeds differ as well. Passing no seed also gives independent rows within a batch, since unseeded requests share one batched draw, and the engine seed then fixes that draw across restarts. An unseeded loop thus shares no draw across requests, the property that separates the two per-request rules from the two shared ones in Table~\ref{tab:twobytwo}.

\end{document}